%% file: iclr_2027_arxiv.tex
\documentclass{article}

\usepackage{iclr2027_conference,times}
\iclrfinalcopy
\usepackage[utf8]{inputenc}
\usepackage[T1]{fontenc}
\usepackage{hyperref}
\usepackage{url}
\usepackage{booktabs}
\usepackage{amsfonts}
\usepackage{nicefrac}
\usepackage{microtype}
\usepackage[dvipsnames,table]{xcolor}

\input{preamble}

\renewcommand{\thefootnote}{\fnsymbol{footnote}}

\title{Compact Motion Embeddings for Feed-forward Unified Human Animation}
\author{
Seonwook Park$^1$ \quad
Hyunsoo Cha$^2$\footnotemark[1] \quad
Amrita Mazumdar$^1$ \quad
Shengze Wang$^1$ \quad
\\ \vspace*{4mm} \bf \hspace{-0.9mm} %
Tianye Li$^1$ \quad
Hanbyul Joo$^2$ \quad
Koki Nagano$^1$ \quad
Shalini De Mello$^1$
\\ \vspace*{2mm}
NVIDIA$^1$ \qquad Seoul National University$^2$
}

\begin{document}
\maketitle

\footnotetext[1]{Part of the work was done during an internship at NVIDIA.}
\renewcommand{\thefootnote}{\arabic{footnote}} 

\lhead{}

\input{section/0_abstract}

\input{section/1_intro}
\input{section/2_related_work}
\input{section/3_method}
\input{section/4_experiments}
\input{section/5_conclusion}

\subsection*{Ethics statement}
Animating a person's image with another person's motion can be misused to create deceptive media (deepfakes); use of such systems should require consent from the people depicted.
Our training and evaluation data come from public video sources and public benchmarks; we use them for research only.

\subsection*{Reproducibility statement}
\autoref{sec:suppl_impl} gives the architecture, parameter counts, losses, loss weights, and training settings.
\autoref{sec:suppl_eval} defines every metric and evaluation protocol.
\autoref{sec:suppl_data} describes the data sources and processing pipeline, and \autoref{sec:suppl_efficiency} the timing protocol.

\subsection*{AI use statement}
We used LLM-based assistants for coding (including parts of the model, evaluation, and benchmarking code), for feedback on experimental design and the interpretation of results, for literature search, and for drafting and editing text.
The authors reviewed and verified all AI-assisted code, analyses, and text, and take full responsibility for the content of this paper.

\bibliographystyle{iclr2027_conference}
\bibliography{main}

\clearpage
\input{section/X_suppl}

\end{document}

%% file: preamble.tex
\usepackage{graphicx}       %
\usepackage{subcaption}     %
\usepackage{amsmath,amssymb}%
\usepackage{mathtools}      %
\usepackage{multirow}       %
\usepackage{xspace}         %
\usepackage[inline]{enumitem}       %
\usepackage[capitalise,noabbrev]{cleveref}  %
\usepackage{algorithm}
\usepackage{algorithmic}
\usepackage{array}
\usepackage{wrapfig}

\graphicspath{{figure/}}
\ifdefined\pdfsuppressptexinfo \pdfsuppressptexinfo=-1 \fi

\newcommand{\ours}{FUHA\xspace}

\usepackage{suffix}
\newcommand\parahead[1]{\vspace{1.5mm}\noindent\textbf{#1.}\medspace}
\WithSuffix\newcommand\parahead*[1]{\vspace{2mm}\noindent\textbf{#1}\medspace}

\definecolor{rankbest}{RGB}{255, 198, 198}
\definecolor{ranksecond}{RGB}{255, 220, 175}
\definecolor{rankthird}{RGB}{255, 240, 175}

\newcolumntype{L}[1]{>{\raggedright\let\newline\\\arraybackslash\hspace{0pt}}m{#1}}
\newcolumntype{C}[1]{>{\centering\let\newline\\\arraybackslash\hspace{0pt}}m{#1}}
\newcolumntype{R}[1]{>{\raggedleft\let\newline\\\arraybackslash\hspace{0pt}}m{#1}}

%% file: section/0_abstract.tex
\begin{abstract}
Animating a reference image with a driver's motion requires a motion representation that captures face, hand, and body articulation but excludes the driver's appearance.
Most body animation methods take motion from external pose estimators and render it with many denoising steps; compact learned motion representations mostly cover the face.
We present \ours, a real-time-capable feed-forward transformer that learns a compact motion embedding of the whole human and uses it for face, upper-body, and full-body animation.
A Q-Former compresses the driver's features from a frozen visual encoder into this embedding.
To disentangle motion from appearance, a novel contrastive loss pulls it toward the motion embedding of the driver's rendered keypoint image and away from that of the same person in another pose.
At inference time, \ours needs no external pose estimators, keypoints, or parametric body models.
As a retrieval key, the motion embedding matches body pose better than the learned global embeddings of ten vision encoders.
Its same-person retrieval rate is 0.1\%, against 9.5\% for its own DINOv3 backbone.
On our upper-body benchmark, \ours outperforms all baselines on every self-reenactment metric and on cross-reenactment identity similarity.
It is competitive on face and full-body and runs at up to 38.9 FPS at $512^2$ on a single RTX 4090.

\end{abstract}

%% file: section/1_intro.tex
\section{Introduction}
\label{sec:intro}

Animating an image of a human (the reference) with the motion of another human (the driver) is a long-standing problem underlying interactive avatars, telepresence, and content creation.
It requires capturing the driver's expressions, limb articulation, and hand pose independently of their identity and appearance, and rendering the reference faithfully under that motion.
Interactive audio-visual avatars add three requirements: (a) region flexibility beyond the face, to the hands, upper body, and full body; (b) real-time capability, for smooth frame rates; and (c) a compact motion representation, since motion updates every frame and must share the per-frame token budget alongside audio and speech, whereas appearance is transmitted once at session start.

\begin{figure*}[t]
  \vskip -2mm
  \centering
  \begin{subfigure}[c]{0.42\linewidth}
    \centering
    \input{table/intro_axes}
    \subcaption{Method categories}
    \label{fig:overview_axes}
  \end{subfigure}\hfill
  \begin{subfigure}[c]{0.55\linewidth}
    \centering
    \includegraphics[width=\linewidth]{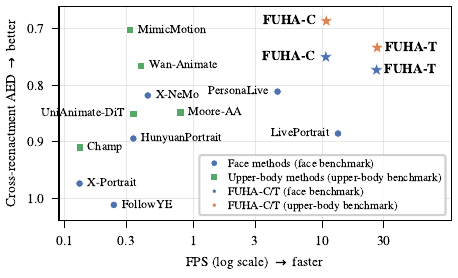}
    \subcaption{Speed vs.\ expression error}
    \label{fig:overview_pareto}
  \end{subfigure}
  \caption{
    \textbf{\ours} drives face, upper-body, and full-body animation from one compact motion embedding.
    (\subref{fig:overview_axes})~Method categories against the three requirements; no prior category meets all three.
    (\subref{fig:overview_pareto})~Speed vs.\ cross-reenactment expression error per benchmark; FUHA-C/T (Cosmos/TAESD3 tokenizers) appear in both.
  }
  \label{fig:overview}
  \vskip -4mm
\end{figure*}

We summarize the existing landscape in \autoref{fig:overview_axes}, where each method category falls short on at least one of these three requirements.
Diffusion-based multi-region methods such as Wan-Animate~\citep{WanAnimate} and UniAnimate-DiT~\citep{UniAnimateDiT} attain high photorealism across face and upper body, but their iterative denoising, typically conditioned on externally estimated keypoints, requires minutes to render seconds of video.
While such models can be distilled to shorten the runtime, they still take body motion from externally estimated keypoints rather than from a compact learned representation.
X-UniMotion~\citep{XUniMotion} and IM-Animation~\citep{IMAnimation} do learn compact whole-body motion embeddings, but decode them with multi-step video diffusion.
At the opposite extreme are face-only animation methods.
LivePortrait~\citep{LivePortrait} and MegaPortraits~\citep{MegaPortraits} provide compact face-only motion embeddings that have, in turn, become a standard input for recent autoregressive real-time motion generators such as ARIG~\citep{ARIG} and Teller~\citep{Teller}.
PersonaLive~\citep{PersonaLive} instead pursues real-time inference by distilling a diffusion-based portrait animator.
None of these categories meets all three requirements: region flexibility, real-time capability, and a compact motion representation.

In this paper we present \textbf{\ours}: \textbf{F}eed-forward \textbf{U}nified \textbf{H}uman \textbf{A}nimation, a single feed-forward model built around a compact motion embedding of the whole human; it is real-time-capable and handles face, upper-body, and full-body regions flexibly.
\ours\ extracts patch features from the driver image via a frozen visual encoder~\citep{DINOv3} and compresses them through a small Q-Former into this embedding.
A similar process happens with appearance information, but with less compression, and the appearance and motion embeddings are used together in a final transformer decoder to render the target image directly.
The result is a network that produces an animated reference frame in a single forward pass (face, hands, and body alike) without iterative denoising, external pose/keypoint estimation, or parametric models of the human (face, hands, or body).
At inference time, \ours takes only a reference image and a driver frame without requiring warm-up frames, per-clip alignment, or other preprocessing routines.
We evaluate two variants (\autoref{fig:overview_pareto}) that differ only in the frozen tokenizer, which encodes the reference image and decodes the output image: Cosmos for FUHA-C and the faster TAESD3 for FUHA-T. At $512^2$ with TorchScript and FP16 on a single RTX 4090, they run at 12.4 and 38.9 FPS, respectively.

Learning this embedding hinges on \textbf{appearance--motion disentanglement}: training reconstructs the driver from a reference of the same person, so appearance can leak into the embedding and resurface as identity errors in cross-reenactment.
We constrain the embedding in three ways.
First, its input: an \textbf{asymmetric encoder pathway} gives tokenizer latents to the appearance branch only, and the motion branch sees frozen visual features alone.
Second, its capacity: the motion embedding is an \textbf{information bottleneck} by design, sized for per-frame motion rather than pixel-level appearance.
Third, its training target: a \textbf{novel contrastive loss} pulls the motion embedding toward that of a rendered keypoint image of the driver and away from that of the reference (same identity, different pose), without requiring explicit keypoint regression.

\parahead{Contributions}
Our contributions are
\begin{enumerate*}[label=(\roman*)]
\item a compact motion embedding of the whole human (face, hands, and body), learned by constraining its input, capacity, and training target, that needs no pose estimator at inference time,
\item \ours, a feed-forward transformer that uses this embedding for face, upper-body, and full-body animation at up to 38.9\,FPS at $512\times 512$ on a single RTX 4090; it outperforms every baseline on all nine self-reenactment metrics of our upper-body benchmark and is competitive on face and full-body, and
\item evidence that the embedding carries motion with little identity: in upper-body cross-reenactment, \ours preserves the reference identity better than every baseline, and in retrieval the embedding returns the query's own person in 0.1\% of queries, against 9.5\% for the learned global embedding of its own DINOv3 backbone (\autoref{tab:retrieval_main}).
\end{enumerate*}

%% file: table/intro_axes.tex
\centering
\scriptsize
\setlength{\tabcolsep}{2pt}
\renewcommand{\arraystretch}{1.0}
\begin{tabular}{@{}L{0.54\linewidth}C{0.13\linewidth}C{0.12\linewidth}C{0.16\linewidth}@{}}
  \toprule
  & {Multi\newline[-0.5mm]Region} & {Real\newline[-0.5mm]Time} & {Compact\newline[-0.5mm] Motion} \\
  \midrule
  \textbf{Multi-region diffusion}\newline[-0.5mm] {\tiny \emph{(e.g., Wan-Animate, UniAnimate-DiT)}}
  & \checkmark & $\times$ & $\times$ \\
  \textbf{Multi-region learned-motion diffusion}\newline[-0.5mm] {\tiny \emph{(e.g., X-UniMotion, IM-Animation)}}
  & \checkmark & $\times$ & \checkmark \\
  \textbf{Face-only feed-forward}\newline[-0.5mm] {\tiny \emph{(e.g., LivePortrait, MegaPortraits)}}
  & $\times$ & \checkmark & \checkmark \\
  \textbf{Face-only diffusion}\newline[-0.5mm] {\tiny \emph{(e.g., X-Portrait, X-NeMo)}}
  & $\times$ & $\times$ & \checkmark/$\times$ \\
  \textbf{Face-only faster diffusion}\newline[-0.5mm] {\tiny \emph{(e.g., PersonaLive)}}
  & $\times$ & \checkmark & $\times$ \\
  \midrule
  \textbf{\ours}~(ours) & \checkmark & \checkmark & \checkmark \\
  \bottomrule
\end{tabular}

%% file: section/2_related_work.tex
\section{Related Work}
\label{sec:related}

\parahead{Portrait Animation}
Portrait animation follows two rendering paradigms.
The first drives a feed-forward decoder with a learned motion embedding: LivePortrait~\citep{LivePortrait}, MegaPortraits~\citep{MegaPortraits}, and VOODOO~XP~\citep{VOODOOXP} render the reference in a single forward pass (other work used parametric 3DMM~\citep{Pirenderer} or StyleGAN~\citep{StyleHeat} intermediaries).
The second conditions a diffusion model on the driving signal and denoises over many steps: X-NeMo~\citep{XNeMo}, X-Portrait~\citep{XPortrait}, Follow-Your-Emoji~\citep{FollowYE}, and HunyuanPortrait~\citep{HunyuanPortrait} trade inference speed for quality; PersonaLive~\citep{PersonaLive} reduces the number of denoising steps through appearance distillation, and FlashPortrait~\citep{FlashPortrait} accelerates sampling with adaptive latent prediction.
Both paradigms target face-centered crops.
\ours keeps the feed-forward design of the first paradigm but extends it beyond the face.

\parahead{Multi-Region Human Animation}
Upper- and full-body animation has mostly relied on externally estimated pose (2D body keypoints, 3D skeletons, SMPL~\citep{SMPL} parameters, or hand keypoints) driving a diffusion backbone.
U-Net-based methods such as Champ~\citep{Champ}, Animate Anyone~\citep{AnimateAnyone}, MimicMotion~\citep{MimicMotion}, UniAnimate~\citep{UniAnimate}, Animate-X~\citep{AnimateX}, and Animate Anyone 2~\citep{AnimateAnyone2} were followed by DiT-based ones (UniAnimate-DiT~\citep{UniAnimateDiT}, Wan-Animate~\citep{WanAnimate}) that improve visual quality at greater computational cost.
OmniHuman-1~\citep{OmniHuman}, DreamActor-M1~\citep{DreamActorM1}, and CyberHost~\citep{CyberHost} combine audio or video with pose in a single diffusion model, but remain multi-step and still extract pose externally.
DiT-based methods rarely run in real time; StreamAvatar~\citep{StreamAvatar}, an audio-driven avatar, does so through autoregressive distillation.
\ours needs no pose estimator at inference time and renders each frame in a single feed-forward pass.

\parahead{Compact Motion Embeddings}
FOMM~\citep{FOMM} and face-vid2vid~\citep{TalkingHead1KH} showed that a low-dimensional motion descriptor can drive a feed-forward warping decoder, and LIA~\citep{LIA} replaced keypoints with orthogonal latent motion directions.
Such embeddings now serve as interfaces for downstream generation: real-time audio-driven head generators~\citep{ARIG,Teller} autoregressively predict LivePortrait-style motion embeddings (implicit keypoints).
X-NeMo~\citep{XNeMo} distills a 1D identity-agnostic motion embedding that conditions a diffusion backbone.
Beyond the face, X-UniMotion~\citep{XUniMotion} learns region-specific motion embeddings (body, face, two hands) jointly with a DiT generator, and IM-Animation~\citep{IMAnimation} compresses per-frame motion into a compact 1D embedding; both decode with multi-step video diffusion.
Pretrained vision encoders~\citep{DINOv3,SigLIP2,Sapiens,Sapiens2,CRADIOv4} also yield learned global embeddings, but these encode identity along with pose (\cref{sec:motion_retrieval}).
\ours learns a compact motion embedding for face, hands, and body that carries little identity, and decodes it with a real-time-capable feed-forward transformer.

%% file: section/3_method.tex
\section{Method}
\label{sec:method}

\begin{figure*}[!ht]
  \centering
  \vskip -2mm
  \includegraphics[width=\linewidth]{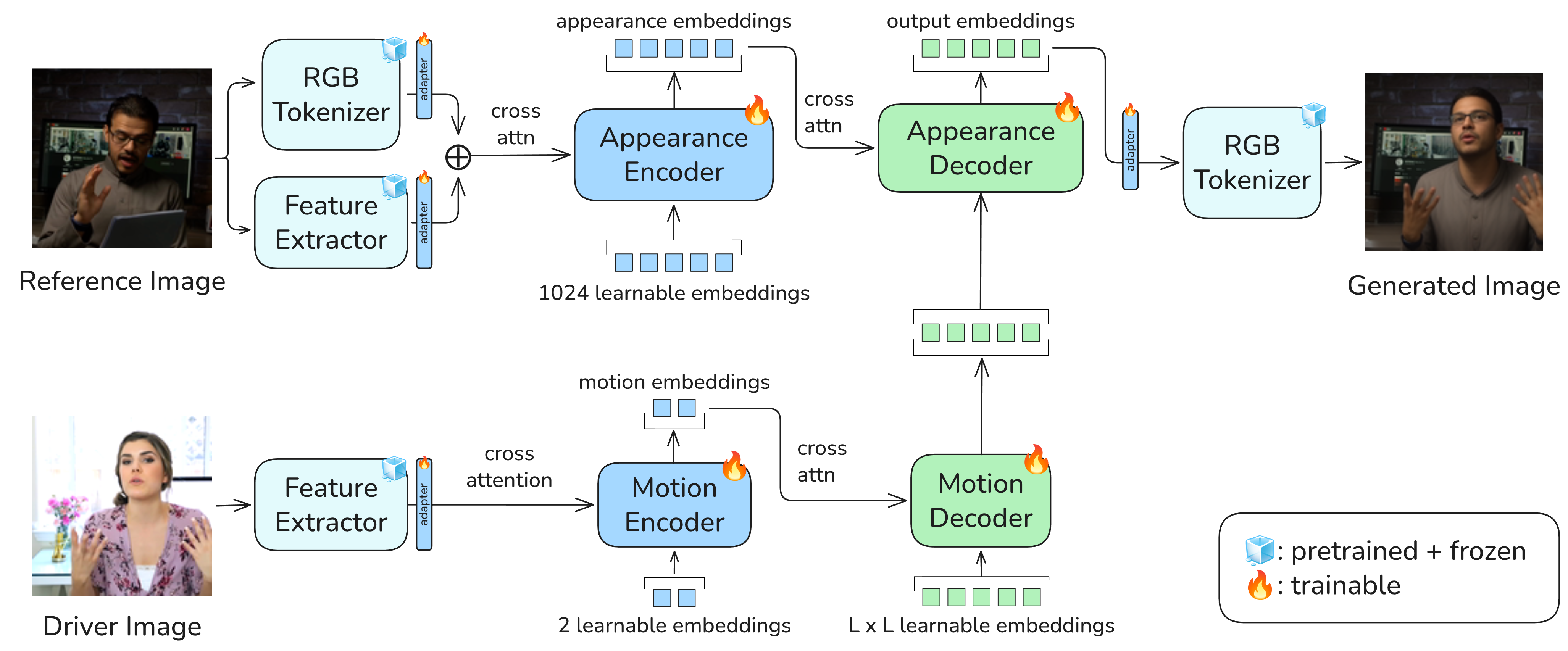}
  \vskip -1mm
  \caption{\textbf{\ours\ inference flow.} Given a reference image and a driver frame, two pretrained, frozen backbones (an RGB tokenizer and a visual feature extractor) produce the cross-attention inputs for two trainable Q-Formers. The Appearance Encoder sums RGB tokens and features and compresses them via cross-attention into a fixed set of appearance embeddings; the Motion Encoder attends to visual features only, yielding a compact motion embedding. A Motion Decoder expands it into a motion map, and the Appearance Decoder cross-attends motion to appearance to produce the final output embeddings, which the frozen RGB tokenizer decodes back to an image. The entire pipeline is feed-forward, frame-by-frame, and real-time-capable.
  }
  \label{fig:architecture}
  \vskip -4mm
\end{figure*}

\subsection{Overview}
\label{sec:overview}

A feed-forward architecture for cross-person reenactment must disentangle \emph{appearance} factors from \emph{motion} factors.
Here, \emph{motion} has typically referred to the pose and expression of the human head, and we extend it to include finger and limb articulation, and overall body translation and scale.
Throughout, ``motion'' denotes a per-frame property, not inter-frame temporal dynamics.

Such appearance--motion disentanglement is hard to enforce in practice, as training only occurs via self-reenactment, where driver and target come from the same clip of the same person.
In this supervision setting, the network can satisfy pixel reconstruction by simply copying the driver (learning a shortcut), sidestepping the disentanglement entirely.
Earlier works achieve disentanglement through an information bottleneck: learned keypoints~\citep{FOMM}, latent motion directions~\citep{LIA}, or a compact mixture of the two~\citep{LivePortrait,MegaPortraits}.
We constrain the motion embedding's input, capacity, and training target.

\parahead{Input: asymmetric features}
We use a frozen visual encoder whose patch features are semantically coherent: they carry pose, expression, and articulation cues across face, hands, and body without region-specific tuning.
A single encoder therefore serves all three settings, and no parametric face or body model is needed.
The two branches read this encoder differently (\autoref{fig:architecture}): the appearance encoder takes its features summed with RGB tokens from a frozen tokenizer~\citep{COSMOS}, while the motion encoder takes the visual features alone.
Because the motion path never sees the tokenizer latents, the motion embedding can only carry what the frozen visual encoder encodes -- primarily semantic content rather than raw photometric detail.
Concretely, we use DINOv3 ViT-B/16~\citep{DINOv3} as the visual encoder. 

\parahead{Capacity: learned compression}
We adopt the Q-Former~\citep{Blip2} architecture to compress the information contained in a large number of patch tokens into a smaller number of learned query tokens, which form the motion embedding, with the input feature map injected via cross-attention.
The same architecture can be used to expand information, by defining a larger number of learnable query tokens.
We take advantage of this in our motion decoder module, where a Q-Former broadcasts the compact motion embedding onto a motion map matching the RGB tokenizer's latent grid.

\parahead{Target: contrastive objective}
Input and capacity alone do not stop appearance, such as hair shape and length, from reaching the motion embedding through the visual features themselves.
We therefore propose a contrastive loss (\autoref{fig:contrastive}): we draw the driver's face, hand, and body keypoints into an image (pose preserved, appearance removed), using keypoints precomputed during dataset construction (\autoref{sec:suppl_data}), and pull the motion embedding toward that of this image and away from that of the reference (same identity, different pose; \cref{sec:objectives}).

\subsection{Architecture}
\label{sec:architecture}

We describe each block of \autoref{fig:architecture} below.
$I_r$ is the reference image, $I_d$ the driver, $\hat{I}$ the output.

\parahead{Frozen backbones}
Two pretrained networks are kept frozen throughout: an RGB tokenizer ($\mathcal{E}_\text{tok}$, $\mathcal{D}_\text{tok}$, we use Cosmos~\citep{COSMOS} for FUHA-C and TAESD3~\citep{TAESD} for FUHA-T) mapping images to and from an $L \times L$ latent grid, and a visual encoder $\phi_\text{vis}$ producing per-patch features.
Image-space losses backpropagate through $\mathcal{D}_\text{tok}$ without updating its weights.

\parahead{Appearance encoder}
Both backbones run on the reference. Their outputs are projected to a shared embedding dimension $d$ by per-modality MLPs, summed at every spatial location, and added to a 2D sinusoidal positional encoding $\sigma_{2D}$, giving the fused reference features $\mathbf{f}_r$.
The Appearance Encoder $E_a$ is a Q-Former with $N_a = 1024$ learned queries that cross-attend to $\mathbf{f}_r$, producing the appearance embeddings $\mathbf{a} \in \mathbb{R}^{N_a \times d}$.

\parahead{Motion encoder}
For the driver, only the feature extractor $\phi_\text{vis}$ runs; its output is projected to $d$ dimensions and added to $\sigma_{2D}$, forming the driver features $\mathbf{f}_d$.
The Motion Encoder $E_m$ is a Q-Former with $N_m = 2$ learned queries that produces the motion embedding $\mathbf{m} = E_m(\mathbf{f}_d) \in \mathbb{R}^{N_m \times d}$, made of $N_m$ motion tokens.

\parahead{Motion decoder}
The Motion Decoder $D_m$ broadcasts $\mathbf{m}$ onto a motion map matched to the tokenizer's latent grid via a Q-Former whose $L^2$ learned queries (with $\sigma_{2D}$ positional encoding) attend to the motion tokens, producing $\mathbf{M} = D_m(\mathbf{m}) \in \mathbb{R}^{L^2 \times d}$.

\parahead{Appearance decoder}
The main decoder $D_a$ is a transformer decoder whose queries are the motion map (with $\sigma_{2D}$ re-applied).
The appearance embeddings $\mathbf{a}$ are injected via vanilla cross-attention layers.
A small MLP $g$ maps its output to the tokenizer's latent dimension, and the frozen tokenizer reconstructs the image: $\hat{I} = \mathcal{D}_\text{tok}\bigl(g\bigl(D_a(\mathbf{M}, \mathbf{a})\bigr)\bigr)$.

The appearance embeddings are computed once per reference image; each driver frame then runs independently through $\phi_\text{vis}$, $E_m$, $D_m$, $D_a$, $g$, $\mathcal{D}_\text{tok}$ in a single feed-forward pass, with no iterative denoising and no external pose estimation modules.

\subsection{Training objectives}
\label{sec:objectives}

\begin{wrapfigure}{r}{0.45\textwidth}
  \centering
  \includegraphics[width=\linewidth]{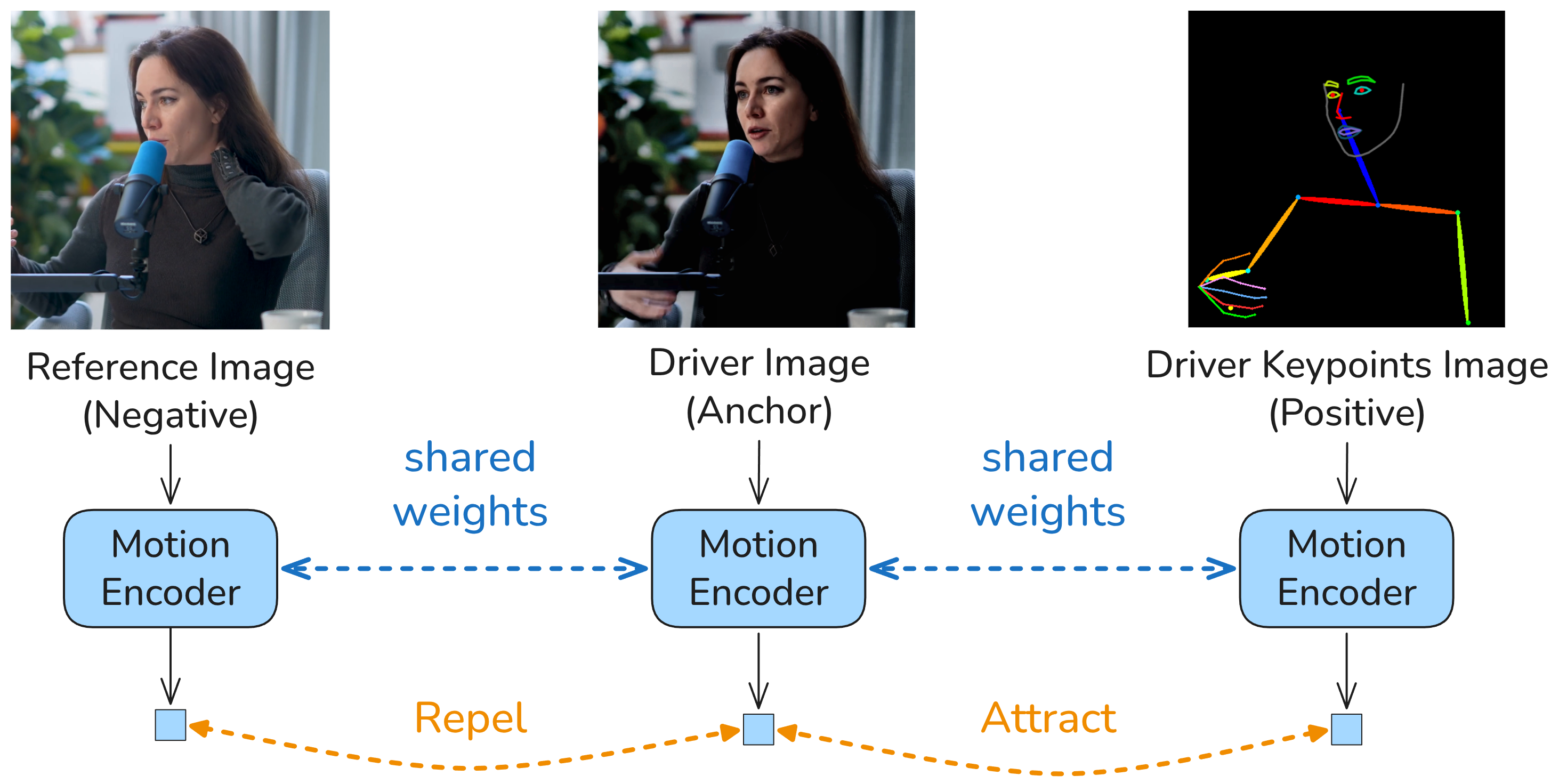}
  \caption{\textbf{Contrastive loss.} The driver's motion embedding is pulled toward the rendered keypoints' and away from the reference's.}
  \label{fig:contrastive}
\end{wrapfigure}

We train \ours\ in a self-reenactment setting: reference and target frames come from the same clip, with the driver being a photometrically augmented version of the target~\citep{XNeMo,VOODOOXP}.
This trains the motion encoder to ignore lighting and color changes that do not affect pose (details in \cref{sec:suppl_training}).
A reconstruction loss $\mathcal{L}_\text{rec}$ combines L1 and LPIPS~\citep{LPIPS}, with spatial reweighting and other details in \autoref{sec:suppl_impl}.
The remaining two losses target appearance--motion disentanglement and identity preservation.

\parahead{Contrastive triplet on the motion embedding}
We further disentangle the motion embedding via a triplet loss on its flattened form $\bar{\mathbf{m}} \in \mathbb{R}^{N_m d}$ (\autoref{fig:contrastive}):
\begin{equation}
    \mathcal{L}_\text{tri} = \operatorname{softplus}\!\bigl( \delta(\bar{\mathbf{m}}_a, \bar{\mathbf{m}}_+) - \delta(\bar{\mathbf{m}}_a, \bar{\mathbf{m}}_-)\bigr),
    \label{eq:triplet}
\end{equation}
where $\delta$ is the cosine distance, the anchor $\bar{\mathbf{m}}_a$ is the driver's flattened motion embedding, the positive $\bar{\mathbf{m}}_+$ is that of a rendered keypoint image of the driver (a foreground crop when face keypoints are missing), and the negative $\bar{\mathbf{m}}_-$ is that of the reference -- same identity, different pose.
The soft-margin form follows \citet{Hermans2017}; the choice of positive and negative is ours.
The positive shares the anchor's pose and the negative its appearance, so the loss ranks pose above appearance (\autoref{sec:suppl_impl}).

\parahead{Identity preservation}
$\mathcal{L}_\text{id}$ is the cosine distance between AdaFace~\citep{AdaFace} embeddings of aligned face crops.
We compare against the driver/target frame (pose-matched and same-identity) to avoid confounding identity drift with pose drift, since AdaFace embeddings are not fully invariant to pose, expression, or shape.

Further training and implementation details, including the full training loss and loss term weights, are in \autoref{sec:suppl_impl}.

%% file: section/4_experiments.tex
\section{Results}
\label{sec:experiments}

\parahead{Implementation Details}
We train FUHA-C and FUHA-T for 60 epochs using a dataset consisting of 1{,}056{,}941 clips from Internet videos, totaling 5{,}172 hours of content.
FUHA-C uses the Cosmos tokenizer, while FUHA-T uses the smaller TAESD3 tokenizer for faster inference. Both use the same transformer architecture and training objective.
Further information about our training and dataset curation process can be found in Appendices~\ref{sec:suppl_impl} and~\ref{sec:suppl_data}, respectively.

\subsection{Experimental Setup}

\ours handles face, upper-body, and full-body animation with a single model and no region-specific fine-tuning. We compare it against baselines in all three settings, with reference images matched to each.

\parahead{Datasets}
We evaluate face self-reenactment on the TalkingHead-1KH test set~\citep{TalkingHead1KH} and extend the LivePortrait~\citep{LivePortrait} protocol for cross-reenactment by using the first 100 FFHQ~\citep{StyleGAN} images as references paired with random TalkingHead clips, 128 frames per clip.
For upper-body, no publicly available standardized evaluation datasets exist, so we curate a new test set from public YouTube videos outside the training split: 96 clips, 10--20\,s long at 25\,Hz. Cross-reenactment uses each video's first frame paired with a different driving clip (96 pairs).
For full-body, we use the TikTok dance test set~\citep{jafarian2021learning}.
All methods are evaluated at $512^2$ image resolution for face and upper-body, and at source image resolution for full-body, where possible, and otherwise with letterboxing and downscaling (due to varying aspect ratios).

\parahead{Baselines}
We select recent baselines with public inference code and pre-trained weights (X-UniMotion and IM-Animation, the closest learned-motion methods, have not released their animation models), and follow each baseline's prescribed preprocessing protocols for image cropping and driving signal construction (e.g., pose estimation and normal map rendering in Champ).
For face animation, we compare against LivePortrait~\citep{LivePortrait}, X-Portrait~\citep{XPortrait}, Follow-Your-Emoji~\citep{FollowYE}, X-NeMo~\citep{XNeMo}, HunyuanPortrait~\citep{HunyuanPortrait}, and PersonaLive~\citep{PersonaLive}.
For upper-body animation, we compare against Moore-AnimateAnyone~\citep{AnimateAnyone}, Champ~\citep{Champ}, MimicMotion~\citep{MimicMotion}, Wan-Animate~\citep{WanAnimate}, and UniAnimate-DiT~\citep{UniAnimateDiT}.
For full-body animation, in addition to the upper-body baselines, we also compare against Animate-X~\citep{AnimateX}.
We omit Animate-X from the upper-body evaluation: unlike the other baselines, it does not condition on precise keypoints, so its output is not pixel-aligned to the driver pose.

\parahead{Evaluation Metrics}
We adopt standard metrics across image fidelity, motion transfer, and identity.
For image fidelity, we report image-level (L1, SSIM, PSNR, LPIPS) and distribution-level (FID~\citep{parmar2021cleanfid}, FVD~\citep{skorokhodov2022stylegan,unterthiner2018towards}) metrics; in cross-reenactment, FID and FVD use the driving videos as the real set.
For motion transfer, we report AED (Average Expression Distance) and APD (Average Pose Distance) using SMIRK~\citep{SMIRK} as a face encoder.
We also adopt the AKD (Average Keypoint Distance) metric~\citep{FOMM,CyberHost,OmniHuman} and report it separately for the body (AKD-B, 17 COCO keypoints) and the hands (AKD-H, 42 keypoints).
For identity, we report ArcFace~\citep{ArcFace} cosine similarity (ID).
Paired metrics such as L1, SSIM, and LPIPS require known targets, so they can only be reported for self-reenactment.
AKD-B and AKD-H are reported only for upper-body and full-body, where the body and hands are visible.
Further details regarding the evaluation metrics are in \autoref{sec:suppl_eval}.

\subsection{Comparison with Baselines}

\input{table/comparison_upperbody_sep26}
\begin{figure*}[!t]
\centering
\setlength{\lineskip}{0pt}\setlength{\parskip}{0pt}
\includegraphics[width=\textwidth]{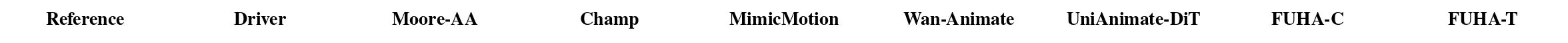}\\[0mm]
\includegraphics[width=\textwidth]{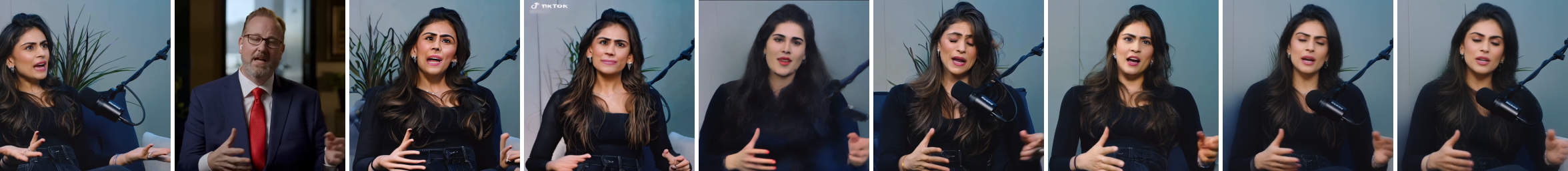}\\[0.5mm]
\includegraphics[width=\textwidth]{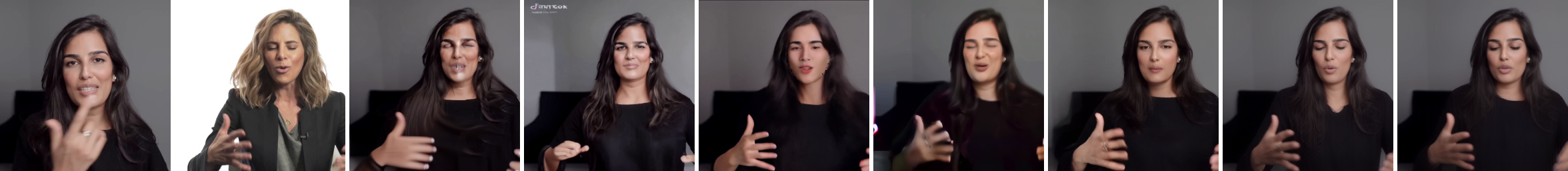}\\[0.5mm]
\includegraphics[width=\textwidth]{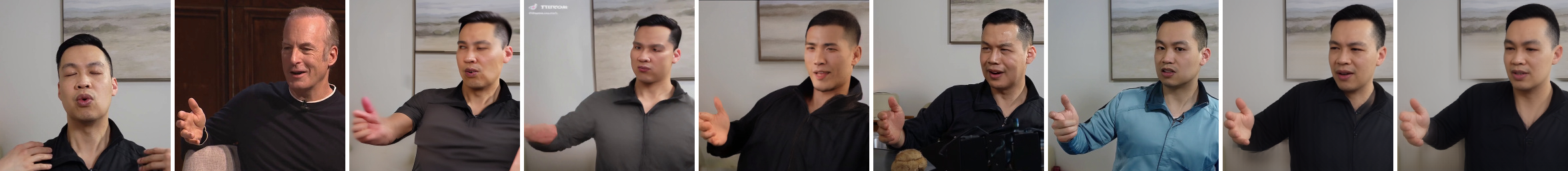}\\[0.5mm]
\includegraphics[width=\textwidth]{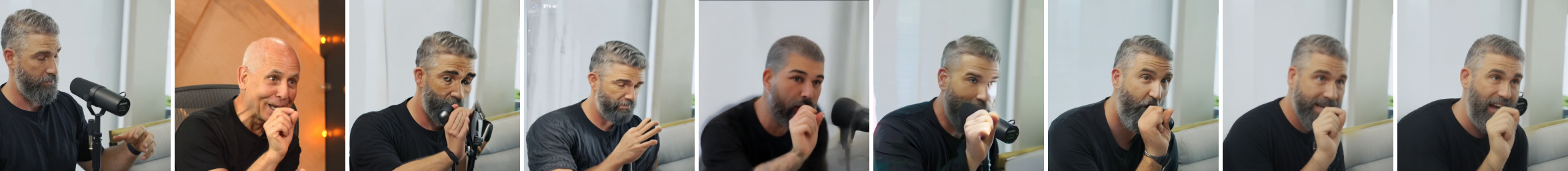}\\[0.5mm]
\vskip -2mm
\caption{\textbf{Qualitative results} for upper-body cross-reenactment. \ours keeps the reference identity while following the driver's expression, body pose, and hand pose.
The final two columns show FUHA-C and FUHA-T, respectively.
More results, including face and full-body, are in \autoref{sec:suppl_qual}.
}
\label{fig:ubody_qual}
\end{figure*}
\parahead{Upper-body Animation}
In \autoref{tab:upperbody}, FUHA-C achieves the best scores across all reported self-reenactment metrics, including image fidelity, identity preservation, and motion transfer.
For cross-reenactment, FUHA-T has the highest ID and FUHA-C the lowest AED.
No baseline has significantly lower cross-reenactment motion errors than FUHA-C (clip-level 95\% intervals; \autoref{sec:suppl_eval}).
UniAnimate-DiT has lower dataset-level FID and FVD, and Moore-AA lower FID.
Rendering FUHA-C's frozen motion embedding with a video diffusion decoder (FUHA-W, Appendix~\ref{sec:suppl_diffusion}) lowers cross-FVD from 171.7 to 125.0, at higher motion errors.
FUHA-T, the faster variant, also outperforms every baseline on all nine self-reenactment metrics.

\autoref{fig:ubody_qual} shows representative qualitative comparisons against the upper-body baselines; side-by-side videos are in the supplementary material.
We observe that each method has strengths but makes compromises.
For example, MimicMotion achieves good AED, AKD-B, and AKD-H scores, but visibly loses the reference identity (also reflected in its low ID score).
Wan-Animate and UniAnimate-DiT render fine detail well.
\ours encodes the reference appearance once and renders every frame from it, without conditioning on previous outputs.
\ours has the lowest self-reenactment hand error (AKD-H 48.5 vs.\ 65.6 for the best baseline) without external pose estimators at inference time; in cross-reenactment, no baseline has significantly lower hand error.

\input{table/comparison_face_sep26}
\parahead{Face Animation}
In \autoref{tab:face}, FUHA-C achieves the best self-reenactment scores for seven of the eight reported metrics, with PersonaLive achieving a lower FID.
For cross-reenactment, FUHA-C has the lowest expression and head-pose errors; X-Portrait and FollowYE lead identity similarity but have the highest expression errors, and several baselines obtain lower FID or FVD.
Here FID and FVD compare outputs for FFHQ references with TalkingHead-1KH driving videos; across this domain gap, every method scores 154--165 FID.
FUHA-T remains competitive on reconstruction and motion transfer, with slightly higher cross-ID than FUHA-C.

\input{table/comparison_fullbody_sep26}
\parahead{Full-body Animation}
We follow previous works and evaluate self-reenactment only on the TikTok test set.
In \autoref{tab:fullbody}, FUHA-C achieves the best image-level, identity, expression, and keypoint scores, leading eight of the ten reported metrics.
UniAnimate-DiT and Wan-Animate have the lowest FID and FVD, respectively.
FUHA-T has higher reconstruction and motion errors than FUHA-C, but remains competitive with the baselines.

\subsection{Ablation Study}
\input{table/ablation_exp23_main}

We ablate the most impactful components of FUHA-C on our upper-body set in \autoref{tab:ablation}.
Each ablation variant and the full model are trained for 60 epochs using the Cosmos tokenizer.
Cross-reenactment is a direct test of appearance--motion disentanglement: appearance leakage in the motion embedding leads to identity confusion and visual artifacts when motion is transferred between identities.

\parahead{Number of motion tokens}
The bottleneck has a sweet spot.
One token is too few: outputs lose identity even in self-reenactment (self-ID 0.134, cross-ID 0.088).
Four tokens keep identity (cross-ID 0.733) but worsen cross-AED (0.686 to 0.761) and every self-reenactment metric.
We use two.

\parahead{Contrastive loss}
Without it, appearance flows through the motion embedding.
Outputs take the driver's hair, gender, and face shape at low detail, but the reference's background and clothing color (\autoref{fig:ablation_qual}).
Cross-ID falls to 0.057, and self-reenactment fails too (self-ID 0.199).
Its lower cross-AED and AKD reflect leaked driver appearance, not better transfer.

\parahead{DINOv3 features in the appearance branch}
DINOv3 features in the appearance branch improve reconstruction and expression transfer.
Removing them increases self-LPIPS from 0.106 to 0.125 and cross-AED from 0.686 to 0.732, though cross-ID improves slightly.

\parahead{Motion decoder}
In this variant, the motion embedding is concatenated with the appearance embeddings, and the appearance decoder cross-attends to both.
Dropping the 2D motion decoder degrades reconstruction and video quality: self-LPIPS increases from 0.106 to 0.116 and cross-FVD rises from 171.7 to 193.0.
Cross-ID also decreases slightly, from 0.701 to 0.689.

\parahead{Identity supervision}
The identity loss helps preserve the reference's appearance.
Removing it reduces self-ID from 0.820 to 0.764 and cross-ID from 0.701 to 0.568.
A ReLU variant of the contrastive loss and qualitative examples are in \autoref{sec:suppl_ablation_full}.

\begin{figure}[t]
\centering
\includegraphics[width=\textwidth]{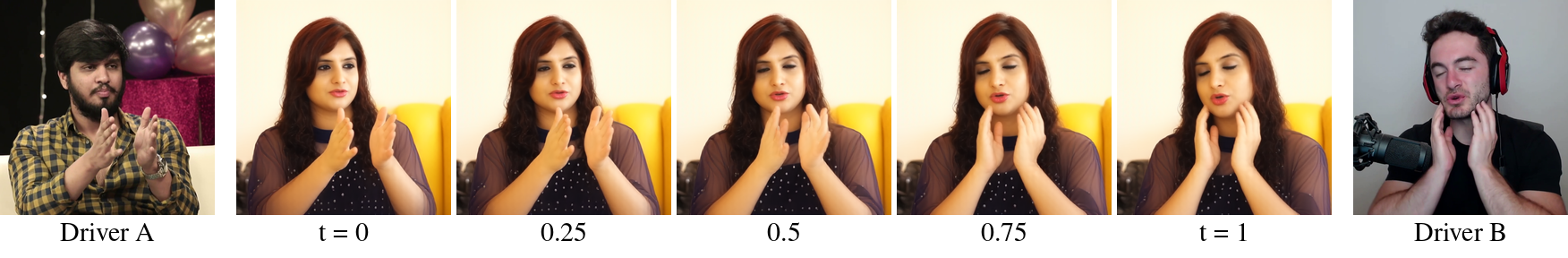}
\caption{\textbf{Motion latent-space walk.} We linearly interpolate between the FUHA-C motion embeddings of two drivers and decode each step on one reference. Over 25 measured pose and expression walks, motion tracks $t$ (median Spearman $\rho = 1.00$, $R^2 = 0.99$) and identity stays with the reference (median ArcFace 0.75; at most 0.19 to either driver; Appendix~\ref{sec:suppl_walks}).}
\label{fig:walk_main}
\end{figure}

\begin{table*}[t]
\begin{minipage}[t]{0.55\textwidth}
\vspace{0pt}
\input{table/efficiency_demo_sep26}
\end{minipage}\hfill
\begin{minipage}[t]{0.43\textwidth}
\vspace{0pt}
\nextfloat
\input{table/retrieval_main}
\end{minipage}
\end{table*}

\subsection{Motion Representation}
\label{sec:motion_retrieval}
We probe the motion embedding by nearest-neighbor retrieval, independently of the animation decoder. Against the learned global embeddings of vision encoders (CLS token or pooled output), FUHA-C has the lowest body-pose error (78.1 pixels) with near-zero same-person retrieval (0.1\%, against 9.5\% for DINOv3; random embeddings give 0.0\%); hand and head-pose results are competitive (\autoref{tab:retrieval_main}). The full protocol, additional baselines, and examples are in Appendix~\ref{sec:suppl_retrieval}. Linear walks between two motion embeddings decode to smooth changes in head pose, expression, and hands while the reference identity holds (\autoref{fig:walk_main}; more in \autoref{fig:latent_walks}). The frozen embedding can also replace the skeleton and face-crop inputs of Wan-Animate, a video diffusion model, with adapters and LoRA fine-tuned on our training data (FUHA-W, Appendix~\ref{sec:suppl_diffusion}).

\subsection{Efficiency}
\label{sec:efficiency}

\autoref{fig:efficiency} compares inference speed and peak VRAM with face and upper-body methods.
On a single RTX Pro 6000, FUHA-C and FUHA-T run at 10.7 and 26.6 FPS, respectively, for upper-body animation at $512^2$. Exported to TorchScript with FP16, they reach 12.4 and 38.9 FPS on a single RTX 4090.
Further timing details are reported in \autoref{sec:suppl_efficiency}.

%% file: table/comparison_upperbody_sep26.tex
\begin{table*}[t]
\centering
\caption{\textbf{Upper-body animation evaluation} on our benchmark at $512^2$
(\colorbox{rankbest}{1st}, \colorbox{ranksecond}{2nd}, \colorbox{rankthird}{3rd}).
}
\vskip -2mm
\label{tab:upperbody}
\setlength{\tabcolsep}{3pt}
\renewcommand{\arraystretch}{1.1}
\scriptsize
\resizebox{\textwidth}{!}{%
\begin{tabular}{lccccccccccccccc}
\toprule
\multirow{2}{*}{Method} & \multicolumn{9}{c}{Self-Reenactment} & \multicolumn{6}{c}{Cross-Reenactment} \\
\cmidrule(lr){2-10} \cmidrule(lr){11-16}
& L1 $\downarrow$ & SSIM $\uparrow$ & LPIPS $\downarrow$
& ID $\uparrow$
& AED $\downarrow$ & AKD-B $\downarrow$ & AKD-H $\downarrow$
& FID $\downarrow$ & FVD $\downarrow$
& ID $\uparrow$
& AED $\downarrow$ & AKD-B $\downarrow$ & AKD-H $\downarrow$
& FID $\downarrow$ & FVD $\downarrow$
\\
\midrule
Moore-AA~\citep{AnimateAnyone} & \cellcolor{rankthird}0.0619 & 0.694 & 0.206 & 0.657 & 0.548 & 42.3 & 81.7 & 26.2 & 185.9 & 0.511 & 0.848 & 56.5 & 103.1 & \cellcolor{ranksecond}53.2 & 368.0 \\
Champ~\citep{Champ} & 0.0980 & 0.611 & 0.335 & 0.571 & 0.670 & 56.3 & 103.2 & 33.8 & 274.8 & 0.452 & 0.910 & 60.2 & 117.8 & 56.4 & 378.8 \\
MimicMotion~\citep{MimicMotion} & 0.0816 & 0.677 & 0.278 & 0.582 & 0.506 & 43.2 & \cellcolor{rankthird}65.6 & 27.1 & 80.3 & 0.131 & \cellcolor{ranksecond}0.702 & \cellcolor{rankthird}53.8 & \cellcolor{rankbest}86.1 & 64.8 & 203.7 \\
Wan-Animate~\citep{WanAnimate} & 0.0643 & 0.712 & 0.219 & 0.668 & \cellcolor{rankthird}0.467 & \cellcolor{rankthird}37.6 & 70.5 & 24.4 & 133.4 & 0.489 & 0.765 & \cellcolor{rankbest}50.8 & 93.4 & 57.5 & 207.4 \\
UniAnimate-DiT~\citep{UniAnimateDiT} & 0.0717 & \cellcolor{rankthird}0.724 & \cellcolor{rankthird}0.201 & \cellcolor{rankthird}0.694 & 0.618 & 46.6 & 86.6 & \cellcolor{rankthird}20.3 & \cellcolor{rankthird}66.0 & \cellcolor{rankthird}0.630 & 0.850 & 55.7 & 97.2 & \cellcolor{rankbest}50.2 & \cellcolor{rankbest}145.8 \\
\midrule
FUHA-C & \cellcolor{rankbest}0.0287 & \cellcolor{rankbest}0.838 & \cellcolor{rankbest}0.106 & \cellcolor{rankbest}0.820 & \cellcolor{rankbest}0.315 & \cellcolor{rankbest}27.1 & \cellcolor{rankbest}48.5 & \cellcolor{rankbest}15.2 & \cellcolor{rankbest}38.9 & \cellcolor{ranksecond}0.701 & \cellcolor{rankbest}0.686 & \cellcolor{ranksecond}51.5 & \cellcolor{ranksecond}88.8 & 55.9 & \cellcolor{ranksecond}171.7 \\
FUHA-T & \cellcolor{ranksecond}0.0320 & \cellcolor{ranksecond}0.826 & \cellcolor{ranksecond}0.117 & \cellcolor{ranksecond}0.812 & \cellcolor{ranksecond}0.348 & \cellcolor{ranksecond}30.6 & \cellcolor{ranksecond}55.9 & \cellcolor{ranksecond}17.0 & \cellcolor{ranksecond}55.3 & \cellcolor{rankbest}0.723 & \cellcolor{rankthird}0.733 & 56.1 & \cellcolor{rankthird}91.9 & \cellcolor{rankthird}55.5 & \cellcolor{rankthird}193.2 \\
\bottomrule
\end{tabular}%
}
\end{table*}

%% file: table/comparison_face_sep26.tex
\begin{table*}[t]
\centering
\caption{\textbf{Face animation evaluation} on the TalkingHead-1KH~\citep{TalkingHead1KH} test split for self-reenactment.
For cross-reenactment we use 100 FFHQ~\citep{StyleGAN} images as reference, paired randomly with a TalkingHead-1KH test set clip.
All methods receive FFHQ-aligned $512^2$ frames.
Table cells are shaded based on per-column rank: \colorbox{rankbest}{1st}, \colorbox{ranksecond}{2nd}, and \colorbox{rankthird}{3rd}.
}
\vskip -2mm
\label{tab:face}
\setlength{\tabcolsep}{4pt}
\renewcommand{\arraystretch}{1.1}
\scriptsize
\resizebox{\textwidth}{!}{%
\begin{tabular}{lccccccccccccc}
\toprule
\multirow{2}{*}{Method} & \multicolumn{8}{c}{Self-Reenactment} & \multicolumn{5}{c}{Cross-Reenactment} \\
\cmidrule(lr){2-9} \cmidrule(lr){10-14}
& L1 $\downarrow$ & SSIM $\uparrow$ & LPIPS $\downarrow$
& ID $\uparrow$
& AED $\downarrow$ & APD $\downarrow$ & FID $\downarrow$ & FVD $\downarrow$
& ID $\uparrow$
& AED $\downarrow$ & APD $\downarrow$ & FID $\downarrow$ & FVD $\downarrow$
\\
\midrule
LivePortrait~\citep{LivePortrait} & 0.0990 & 0.655 & 0.359 & \cellcolor{rankthird}0.868 & 0.341 & 0.0256 & 40.4 & 224.7 & 0.658 & 0.885 & 0.0705 & 164.9 & \cellcolor{rankbest}533.0 \\
X-Portrait~\citep{XPortrait} & 0.0820 & 0.685 & 0.271 & 0.855 & 0.414 & 0.0449 & 37.1 & 271.3 & \cellcolor{rankbest}0.793 & 0.974 & 0.1186 & 162.9 & 655.5 \\
FollowYE~\citep{FollowYE} & 0.0746 & \cellcolor{rankthird}0.699 & 0.250 & 0.844 & 0.440 & 0.0282 & 39.4 & \cellcolor{rankthird}184.6 & \cellcolor{ranksecond}0.747 & 1.012 & 0.0483 & 160.0 & 678.2 \\
X-NeMo~\citep{XNeMo} & 0.1072 & 0.600 & 0.376 & 0.832 & 0.401 & 0.0206 & 43.8 & 245.4 & 0.654 & 0.818 & 0.0486 & \cellcolor{rankbest}154.3 & \cellcolor{ranksecond}546.6 \\
HunyuanPortrait~\citep{HunyuanPortrait} & 0.0943 & 0.660 & 0.312 & 0.853 & 0.376 & 0.0313 & \cellcolor{ranksecond}35.0 & 203.8 & 0.691 & 0.894 & 0.0880 & 159.8 & 571.8 \\
PersonaLive~\citep{PersonaLive} & \cellcolor{rankthird}0.0721 & 0.698 & \cellcolor{rankthird}0.239 & \cellcolor{ranksecond}0.870 & \cellcolor{ranksecond}0.323 & \cellcolor{rankthird}0.0171 & \cellcolor{rankbest}32.6 & 210.1 & \cellcolor{rankthird}0.698 & \cellcolor{rankthird}0.811 & \cellcolor{rankthird}0.0365 & \cellcolor{ranksecond}154.9 & \cellcolor{rankthird}561.3 \\
\midrule
FUHA-C & \cellcolor{rankbest}0.0526 & \cellcolor{rankbest}0.782 & \cellcolor{rankbest}0.185 & \cellcolor{rankbest}0.876 & \cellcolor{rankbest}0.293 & \cellcolor{rankbest}0.0131 & \cellcolor{ranksecond}35.0 & \cellcolor{rankbest}153.8 & 0.654 & \cellcolor{rankbest}0.750 & \cellcolor{rankbest}0.0321 & 158.2 & 645.3 \\
FUHA-T & \cellcolor{ranksecond}0.0561 & \cellcolor{ranksecond}0.770 & \cellcolor{ranksecond}0.196 & 0.863 & \cellcolor{rankthird}0.334 & \cellcolor{ranksecond}0.0154 & \cellcolor{rankthird}36.9 & \cellcolor{ranksecond}180.7 & 0.661 & \cellcolor{ranksecond}0.773 & \cellcolor{ranksecond}0.0324 & \cellcolor{rankthird}156.5 & 645.1 \\
\bottomrule
\end{tabular}%
}
\end{table*}

%% file: table/comparison_fullbody_sep26.tex
\begin{table*}[t]
\centering
\caption{\textbf{Full-body animation evaluation} on the TikTok~\citep{jafarian2021learning} test set (self-reenactment).
(\colorbox{rankbest}{1st}, \colorbox{ranksecond}{2nd}, \colorbox{rankthird}{3rd}).
}
\vskip -2mm
\label{tab:fullbody}
\setlength{\tabcolsep}{6pt}
\renewcommand{\arraystretch}{1.1}
\small
\resizebox{\textwidth}{!}{%
\begin{tabular}{lcccccccccc}
\toprule
Method
& L1 $\downarrow$ & SSIM $\uparrow$ & PSNR $\uparrow$ & LPIPS $\downarrow$
& ID $\uparrow$
& AED $\downarrow$ & AKD-B $\downarrow$ & AKD-H $\downarrow$
& FID $\downarrow$ & FVD $\downarrow$ \\
\midrule
Moore-AA~\citep{AnimateAnyone} & 0.0812 & 0.770 & 17.58 & 0.302 & 0.448 & 0.733 & 48.6 & 87.5 & 44.8 & 235.8 \\
Champ~\citep{Champ} & 0.1223 & 0.719 & 14.45 & 0.399 & 0.356 & 0.993 & 85.3 & 118.9 & 42.2 & 281.1 \\
MimicMotion~\citep{MimicMotion} & 0.1118 & 0.728 & 15.40 & 0.363 & 0.282 & 0.778 & 65.8 & 88.9 & 37.9 & \cellcolor{rankthird}188.7 \\
Animate-X~\citep{AnimateX} & 0.1145 & 0.745 & 15.00 & 0.385 & 0.421 & 0.961 & 91.3 & 151.2 & \cellcolor{ranksecond}31.0 & 229.5 \\
Wan-Animate~\citep{WanAnimate} & 0.0815 & 0.768 & 17.76 & 0.310 & 0.427 & \cellcolor{rankthird}0.705 & \cellcolor{rankthird}46.8 & 84.5 & \cellcolor{rankthird}36.1 & \cellcolor{rankbest}168.2 \\
UniAnimate-DiT~\citep{UniAnimateDiT} & \cellcolor{rankthird}0.0672 & \cellcolor{rankthird}0.818 & \cellcolor{rankthird}19.11 & \cellcolor{ranksecond}0.247 & \cellcolor{rankthird}0.458 & \cellcolor{rankthird}0.705 & \cellcolor{ranksecond}44.9 & \cellcolor{ranksecond}76.4 & \cellcolor{rankbest}25.9 & \cellcolor{ranksecond}170.8 \\
\midrule
FUHA-C & \cellcolor{rankbest}0.0559 & \cellcolor{rankbest}0.852 & \cellcolor{rankbest}20.50 & \cellcolor{rankbest}0.236 & \cellcolor{rankbest}0.519 & \cellcolor{rankbest}0.605 & \cellcolor{rankbest}44.2 & \cellcolor{rankbest}69.5 & 37.9 & 230.5 \\
FUHA-T & \cellcolor{ranksecond}0.0630 & \cellcolor{ranksecond}0.840 & \cellcolor{ranksecond}19.44 & \cellcolor{rankthird}0.258 & \cellcolor{ranksecond}0.508 & \cellcolor{ranksecond}0.637 & 51.2 & \cellcolor{rankthird}82.1 & 40.0 & 263.7 \\
\bottomrule
\end{tabular}%
}
\vskip -3mm
\end{table*}

%% file: table/ablation_exp23_main.tex
\begin{table*}[t]
\centering
\caption{\textbf{FUHA-C ablations} on the upper-body set at $512^2$.
\cref{tab:ablation_sep26} adds a ReLU variant of the contrastive loss.
(\colorbox{rankbest}{1st}, \colorbox{ranksecond}{2nd}, \colorbox{rankthird}{3rd}).}
\vskip -2mm
\label{tab:ablation}
\setlength{\tabcolsep}{4pt}
\scriptsize
\resizebox{\textwidth}{!}{%
\begin{tabular}{lcccccccccc}
\toprule
& \multicolumn{4}{c}{Self-Reenactment} & \multicolumn{6}{c}{Cross-Reenactment} \\
\cmidrule(lr){2-5} \cmidrule(lr){6-11}
Variant & LPIPS $\downarrow$ & ID $\uparrow$ & FID $\downarrow$ & FVD $\downarrow$
& ID $\uparrow$ & AED $\downarrow$ & AKD-B $\downarrow$ & AKD-H $\downarrow$
& FID $\downarrow$ & FVD $\downarrow$ \\
\midrule
1 motion token & 0.351 & 0.134 & 78.8 & 258.5 & 0.088 & 0.724 & 57.8 & 92.6 & 88.0 & 310.8 \\
2 motion tokens & \cellcolor{rankbest}0.106 & \cellcolor{rankbest}0.820 & \cellcolor{rankbest}15.2 & \cellcolor{rankbest}38.9 & \cellcolor{rankthird}0.701 & 0.686 & \cellcolor{ranksecond}51.5 & 88.8 & \cellcolor{ranksecond}55.9 & \cellcolor{ranksecond}171.7 \\
4 motion tokens & 0.127 & \cellcolor{ranksecond}0.812 & 17.7 & 63.1 & \cellcolor{rankbest}0.733 & 0.761 & 58.9 & 91.6 & \cellcolor{rankbest}55.7 & 201.9 \\
\midrule
w/o contrastive loss & 0.340 & 0.199 & 88.8 & 263.9 & 0.057 & \cellcolor{rankbest}0.606 & \cellcolor{rankbest}47.8 & \cellcolor{rankbest}80.3 & 101.1 & 325.9 \\
w/o DINO in appearance & 0.125 & \cellcolor{rankthird}0.811 & 18.3 & 59.4 & \cellcolor{ranksecond}0.712 & 0.732 & 55.8 & 91.1 & \cellcolor{rankthird}56.1 & 202.4 \\
w/o 2D motion decoder & \cellcolor{rankthird}0.116 & 0.799 & \cellcolor{rankthird}16.9 & \cellcolor{rankthird}49.2 & 0.689 & \cellcolor{rankthird}0.685 & \cellcolor{rankthird}51.9 & \cellcolor{rankthird}86.9 & 57.6 & \cellcolor{rankthird}193.0 \\
w/o ID loss & \cellcolor{ranksecond}0.110 & 0.764 & \cellcolor{ranksecond}15.6 & \cellcolor{ranksecond}41.5 & 0.568 & \cellcolor{ranksecond}0.671 & 52.1 & \cellcolor{ranksecond}86.5 & 56.8 & \cellcolor{rankbest}170.6 \\
FUHA-C & \cellcolor{rankbest}0.106 & \cellcolor{rankbest}0.820 & \cellcolor{rankbest}15.2 & \cellcolor{rankbest}38.9 & \cellcolor{rankthird}0.701 & 0.686 & \cellcolor{ranksecond}51.5 & 88.8 & \cellcolor{ranksecond}55.9 & \cellcolor{ranksecond}171.7 \\
\bottomrule
\end{tabular}%
}
\end{table*}

%% file: table/efficiency_demo_sep26.tex
\centering
\caption{\textbf{Inference efficiency.} FPS and peak VRAM (GB) at $512^2$ on a single GPU.
(\colorbox{rankbest}{1st}, \colorbox{ranksecond}{2nd}, \colorbox{rankthird}{3rd}).}
\label{fig:efficiency}
\vskip -1.5mm
\begin{subtable}[t]{0.48\linewidth}
\centering
\caption{Face methods.}
\vskip -1mm
\setlength{\tabcolsep}{2pt}
\renewcommand{\arraystretch}{1.05}
\scriptsize
\resizebox{\linewidth}{!}{%
\begin{tabular}{lcc}
\toprule
Method & FPS $\uparrow$ & VRAM $\downarrow$ \\
\midrule
LivePortrait & \cellcolor{ranksecond}13.20 & \cellcolor{rankbest}1.4 \\
X-Portrait & 0.13 & 59.1 \\
FollowYE & 0.24 & 8.4 \\
X-NeMo & 0.44 & 20.2 \\
HunyuanPortrait & 0.34 & 22.3 \\
PersonaLive & 4.48 & 9.4 \\
\midrule
FUHA-C & \cellcolor{rankthird}10.62 & \cellcolor{rankthird}3.5 \\
FUHA-T & \cellcolor{rankbest}26.29 & \cellcolor{ranksecond}2.5 \\
\bottomrule
\end{tabular}%
}
\label{fig:efficiency_face}
\end{subtable}
\hfill
\begin{subtable}[t]{0.50\linewidth}
\centering
\caption{Upper-body methods.}
\vskip -1mm
\setlength{\tabcolsep}{2pt}
\renewcommand{\arraystretch}{1.05}
\scriptsize
\resizebox{\linewidth}{!}{%
\begin{tabular}{lcc}
\toprule
Method & FPS $\uparrow$ & VRAM $\downarrow$ \\
\midrule
Moore-AA & \cellcolor{rankthird}0.79 & \cellcolor{rankthird}9.9 \\
Champ & 0.13 & 11.6 \\
MimicMotion & 0.32 & 17.3 \\
Wan-Animate & 0.39 & 54.6 \\
UniAnimate-DiT & 0.34 & 39.8 \\
\midrule
FUHA-C & \cellcolor{ranksecond}10.71 & \cellcolor{ranksecond}4.7 \\
FUHA-T & \cellcolor{rankbest}26.60 & \cellcolor{rankbest}3.7 \\
\bottomrule
\end{tabular}%
}
\label{fig:efficiency_body}
\vskip -2mm
\end{subtable}

%% file: table/retrieval_main.tex
\centering
\caption{\textbf{Motion retrieval.} Pose errors use a person-disjoint gallery; same-person retrieval uses an overlapping gallery. Protocol and full results in Appendix~\ref{sec:suppl_retrieval}.}
\label{tab:retrieval_main}
\setlength{\tabcolsep}{2pt}
\scriptsize
\begin{tabular}{lcccc}
\toprule
Encoder & Body & Hand & Head & Same \\
 & AKD $\downarrow$ & AKD $\downarrow$ & pose $\downarrow$ & person $\downarrow$ \\
 & (px) & (px) & (deg.) & (\%) \\
\midrule
Random & 110.8 & 166.2 & 10.61 & 0.0 \\
Sapiens2 0.1B & 85.5 & 143.8 & 9.09 & 9.5 \\
DINOv3 ViT-B/16 & 84.9 & 142.6 & 9.19 & 9.5 \\
\midrule
FUHA-C & 78.1 & 140.4 & 9.18 & 0.1 \\
\bottomrule
\end{tabular}

%% file: section/5_conclusion.tex
\section{Conclusion}
\label{sec:conclusion}

We present \ours, a real-time-capable feed-forward transformer for face, upper-body, and full-body animation, driven by a compact learned motion embedding of the whole human.
Its input, capacity, and training target are constrained, so it carries little identity: it retrieves the query's own person in only 0.1\% of queries.
Linear walks through the motion embedding decode to smooth, identity-preserving motion, and the frozen embedding can drive a video diffusion model in place of keypoints.
On our upper-body benchmark, \ours leads every self-reenactment metric and outperforms every baseline on cross-reenactment identity; it is competitive on face and full-body, with FUHA-T reaching 38.9 FPS at $512^2$ on a single RTX 4090.
\ours renders each frame independently and without a generative prior, so fine detail is softer than with multi-step diffusion; we discuss limitations and next steps in \autoref{sec:suppl_limitations}.

%% file: section/X_suppl.tex
\appendix
\renewcommand{\sectionautorefname}{Appendix}
\renewcommand{\subsectionautorefname}{Appendix}

\counterwithin{figure}{section}
\counterwithin{table}{section}

\section*{Appendix}

This appendix is organized as follows.
\autoref{sec:suppl_motion} analyzes the motion embedding through nearest-neighbor retrieval, latent-space walks, and a video diffusion decoder (FUHA-W).
Appendices~\ref{sec:suppl_qual} and~\ref{sec:suppl_ablation_full} present temporal-consistency and additional qualitative results, and the full ablation study.
Appendices~\ref{sec:suppl_impl}--\ref{sec:suppl_efficiency} describe the implementation, evaluation, datasets, and timing protocol, and \autoref{sec:suppl_limitations} discusses limitations.

\input{section/retrieval_sep26}

\subsection{Conditioning a Diffusion Model}
\label{sec:suppl_diffusion}

We test whether the motion embedding transfers to a different decoder by conditioning Wan2.2-Animate-14B~\citep{WanAnimate} on the frozen FUHA-C motion embedding and the motion maps decoded from it, denoted FUHA-W.
A convolutional adapter maps FUHA-C's mean-centered motion maps to Wan's 16-channel pose input with $4\times$ temporal downsampling.
We mean-center each motion map by subtracting a fixed per-position mean, computed over 6{,}400 training frames from 200 clips.
A two-layer MLP maps the flattened motion embedding to 512-dimensional expression features, replacing Wan's face-crop motion encoder.
We retain Wan's reference-image conditioning and train the adapters, pose-input convolution, expression encoder, and rank-64 LoRA in its attention and feed-forward layers for 12{,}000 self-reenactment steps on our training data with a flow-matching loss.
FUHA-C, the Wan VAE, and the CLIP image encoder remain frozen; keypoints are used only for face and hand weighting of the training loss.

\autoref{tab:wan} compares FUHA-W with FUHA-C and diffusion baselines across all three driving regions.
For upper-body animation, FUHA-W lowers FVD relative to FUHA-C (self 38.9 to 27.9; cross 171.7 to 125.0), while FID worsens (self 15.2 to 18.4; cross 55.9 to 56.4), ID drops (self 0.820 to 0.781; cross 0.701 to 0.676), and expression and keypoint errors rise.
For face animation, FUHA-W improves cross-ID and cross-FVD but is worse than FUHA-C on every self-reenactment metric and on cross-reenactment expression and head-pose errors.
On full-body animation, FUHA-C performs better across the reported metrics.

Compared with the released Wan-Animate and UniAnimate-DiT, FUHA-W achieves lower upper-body cross-FVD (125.0 versus 207.4 and 145.8) and higher cross-ID (0.676 versus 0.489 and 0.630). Against Wan-Animate, its cross-reenactment motion errors are higher (AED 0.827 versus 0.765, AKD-B 61.4 versus 50.8, AKD-H 94.8 versus 93.4).
For face animation, FUHA-W improves self-FVD over PersonaLive (189.2 versus 210.1) and cross-ID (0.763 versus 0.698), while PersonaLive has lower cross-AED (0.811 versus 0.929) and cross-FVD (561.3 versus 581.4).
The gains do not extend to full-body FVD, where Wan-Animate and UniAnimate-DiT score 168.2 and 170.8, compared with 260.8 for FUHA-W.
With 10 denoising steps at $512^2$, FUHA-W runs at 0.74 FPS on a single RTX Pro 6000, compared with 10.7 FPS for FUHA-C.
The frozen motion embedding thus drives a video diffusion model in place of skeletons and face crops; the diffusion prior improves upper-body video quality at the cost of speed and motion accuracy.

\input{table/wan_dit_suppl}

\begin{figure*}[p]
\centering
\includegraphics[width=\textwidth,height=0.88\textheight,keepaspectratio]{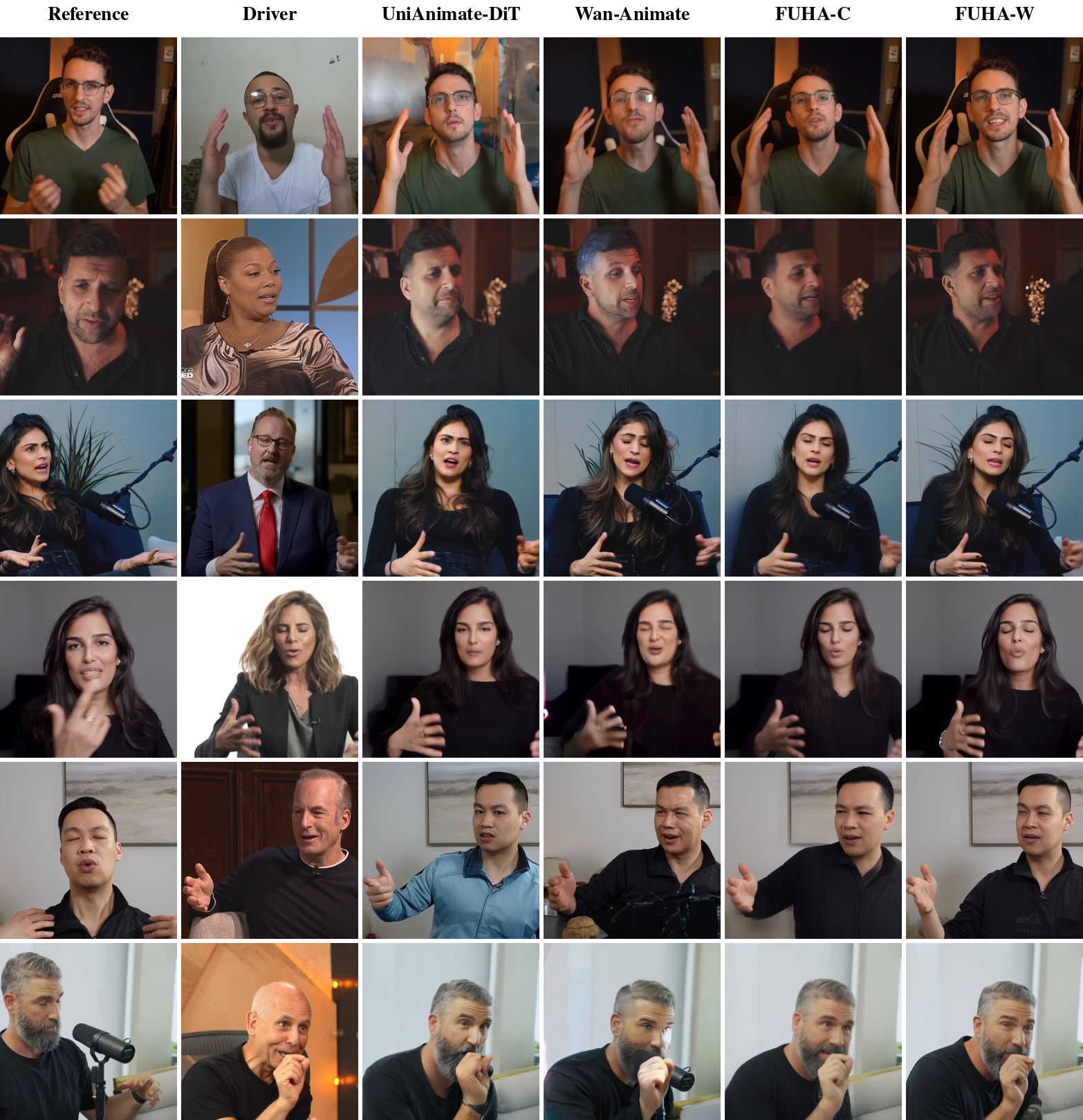}
\caption{\textbf{Qualitative results with FUHA-W.} The diffusion decoder used in FUHA-W yields improved visual details at a slight expense of pose accuracy. All methods use the same reference image and driving frame in each row.}
\label{fig:wan_qual}
\end{figure*}

\autoref{fig:wan_qual} compares FUHA-C and FUHA-W with UniAnimate-DiT and Wan-Animate on upper-body cross-reenactment.

\clearpage

\clearpage

\section{Additional Results}
\label{sec:suppl_qual}
\subsection{Temporal Consistency}
\label{sec:suppl_tlp}
\autoref{tab:tlp} reports tLP~\citep{TecoGAN} for self-reenactment: the mean absolute difference between the LPIPS of consecutive output frames and that of consecutive ground-truth frames.
Cross-reenactment has no ground-truth motion, so tLP is not reported there.
FUHA-C has the lowest tLP on face and upper-body, although it renders every frame independently.
It is trained with reconstruction losses and uses keypoints only as soft guidance through the contrastive loss, which likely keeps its output close to each driver frame, so its frame-to-frame change tracks that of the ground truth.
On full-body, three baselines are lower.

\input{table/tlp_self}

\subsection{Qualitative Comparisons}

\begin{figure*}[!ht]
\centering
\setlength{\lineskip}{0pt}\setlength{\parskip}{0pt}
\includegraphics[width=\textwidth]{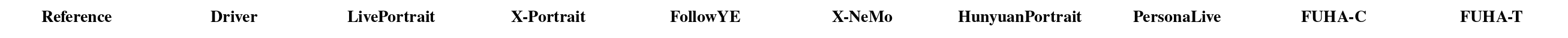}\\[0mm]
\includegraphics[width=\textwidth]{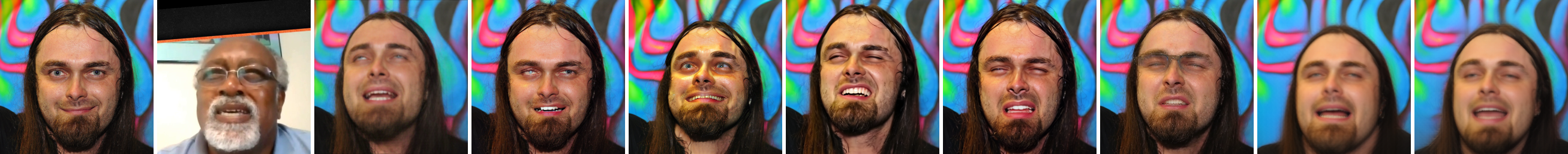}\\[0.5mm]
\includegraphics[width=\textwidth]{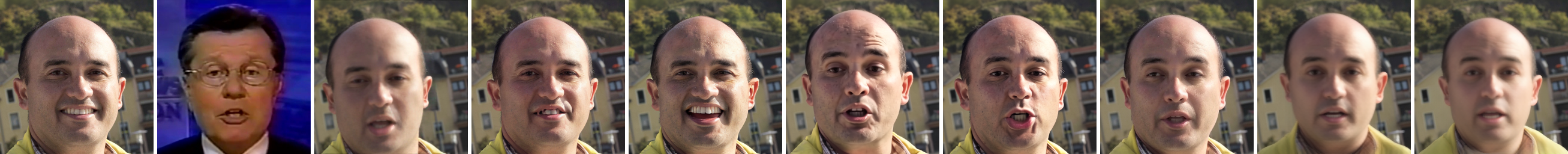}\\[0.5mm]
\includegraphics[width=\textwidth]{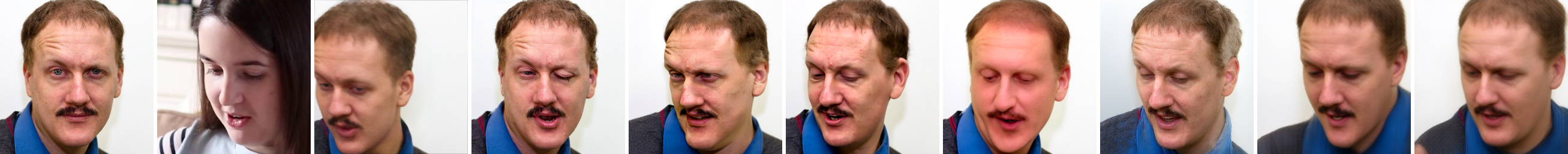}\\[0.5mm]
\includegraphics[width=\textwidth]{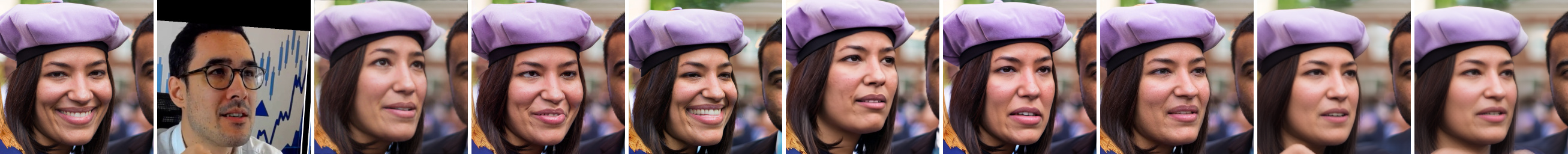}\\[0.5mm]
\includegraphics[width=\textwidth]{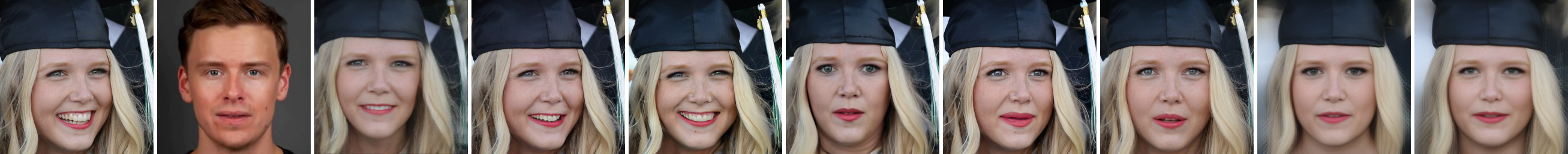}\\[0.5mm]
\includegraphics[width=\textwidth]{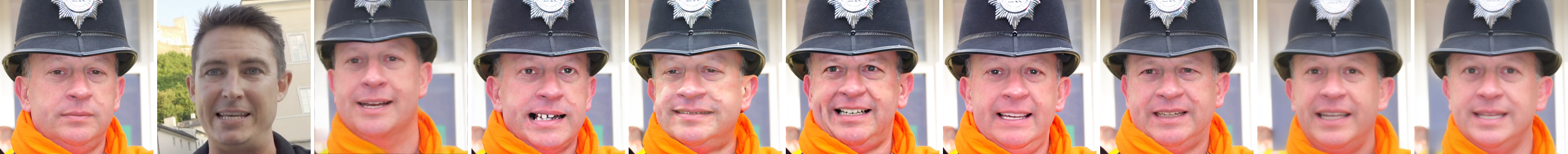}\\[0.5mm]
\caption{\textbf{Qualitative results} for face animation. The final two columns show FUHA-C and FUHA-T, respectively.
Beyond head pose and facial expression, \ours captures shoulder pose and camera-relative head distance.
}
\label{fig:face_qual_suppl}
\end{figure*}

\begin{figure*}[!ht]
\centering
\setlength{\lineskip}{0pt}\setlength{\parskip}{0pt}
\includegraphics[width=\textwidth]{exp23_preview/imgs_ubody/labels.png}\\[0mm]
\includegraphics[width=\textwidth]{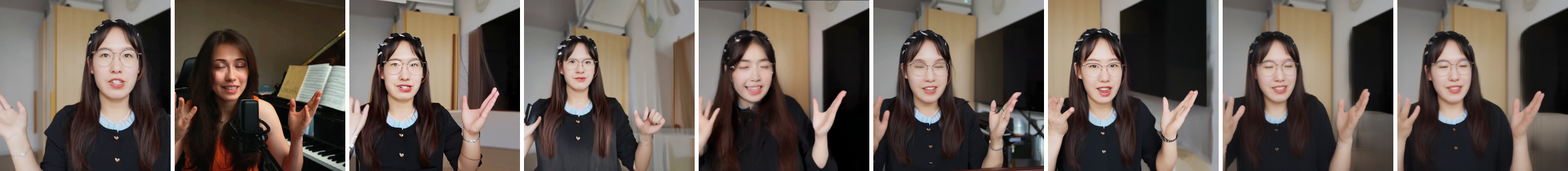}\\[0.5mm]
\includegraphics[width=\textwidth]{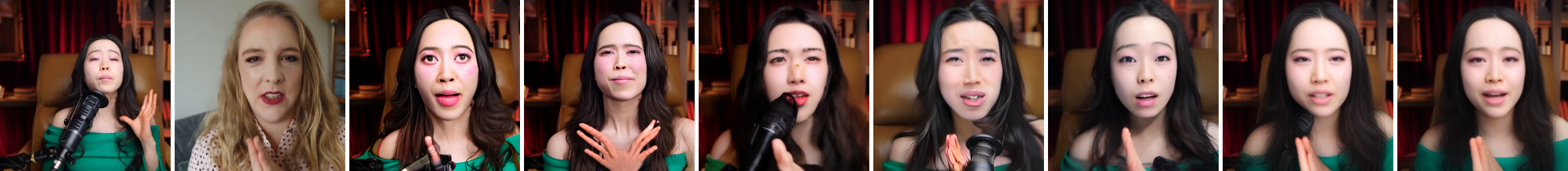}\\[0.5mm]
\includegraphics[width=\textwidth]{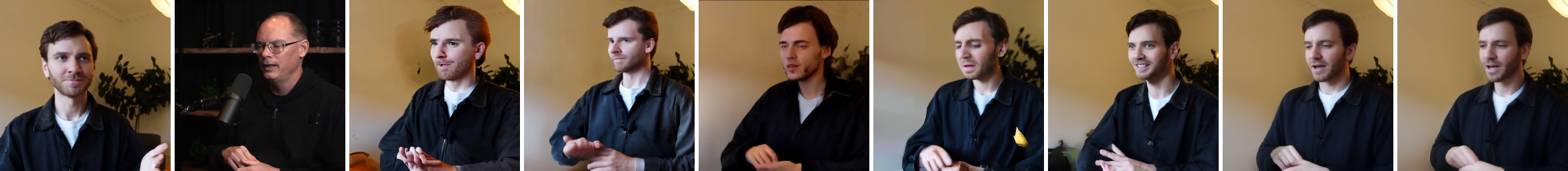}\\[0.5mm]
\includegraphics[width=\textwidth]{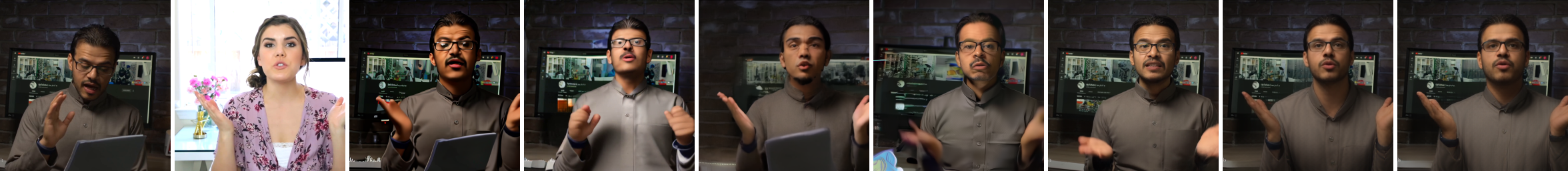}\\[0.5mm]
\includegraphics[width=\textwidth]{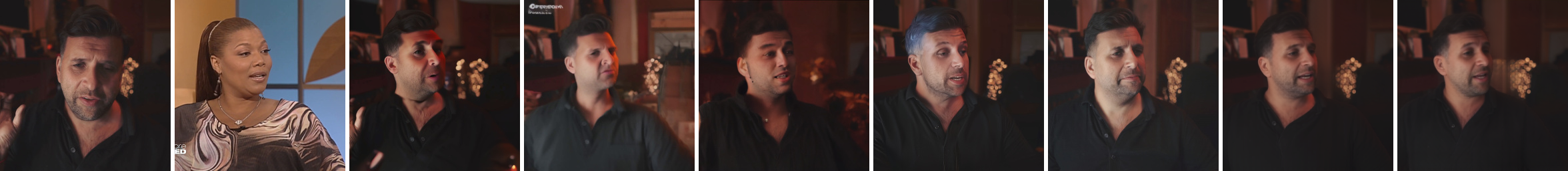}\\[0.5mm]
\caption{\textbf{Additional qualitative results} for upper-body animation. The final two columns show FUHA-C and FUHA-T, respectively.}
\label{fig:ubody_qual_suppl}
\end{figure*}

\begin{figure*}[!ht]
\centering
\setlength{\lineskip}{0pt}\setlength{\parskip}{0pt}
\includegraphics[width=\textwidth]{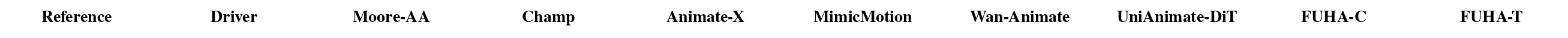}\\[0mm]
\includegraphics[width=\textwidth]{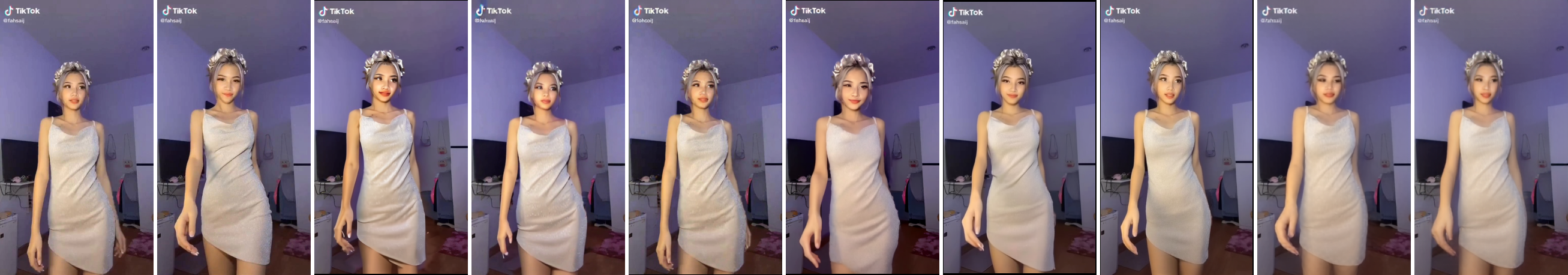}\\[0.5mm]
\includegraphics[width=\textwidth]{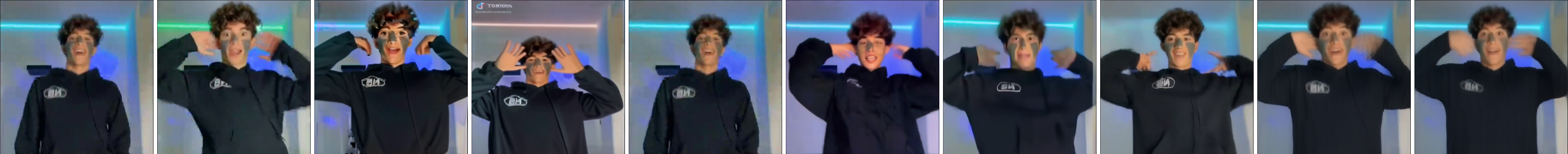}\\[0.5mm]
\includegraphics[width=\textwidth]{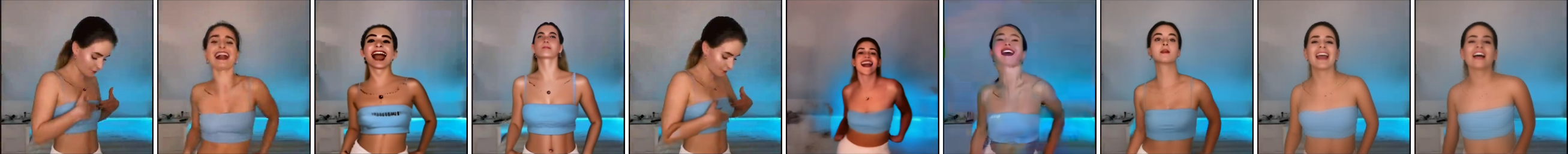}\\[0.5mm]
\includegraphics[width=\textwidth]{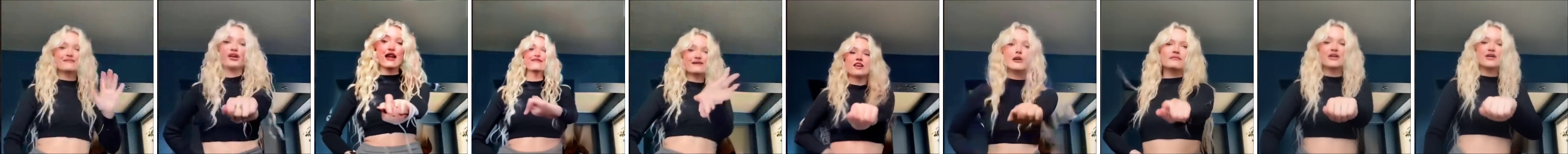}\\[0.5mm]
\caption{\textbf{Qualitative results} for full-body animation. The final two columns show FUHA-C and FUHA-T, respectively.
\ours transfers body pose and facial expression accurately.
}
\label{fig:fbody_qual_suppl}
\end{figure*}

We provide extended qualitative comparisons for face (\autoref{fig:face_qual_suppl}), upper-body (\autoref{fig:ubody_qual_suppl}), and full-body (\autoref{fig:fbody_qual_suppl}) animation against the baselines from \autoref{tab:face}, \autoref{tab:upperbody}, and \autoref{tab:fullbody}, respectively.

\clearpage
\section{Full Ablation Study}
\label{sec:suppl_ablation_full}

\begin{figure*}[!ht]
\centering
\setlength{\lineskip}{0pt}\setlength{\parskip}{0pt}
\includegraphics[width=\textwidth]{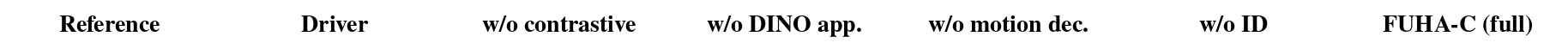}\\[0.5mm]
\includegraphics[width=\textwidth]{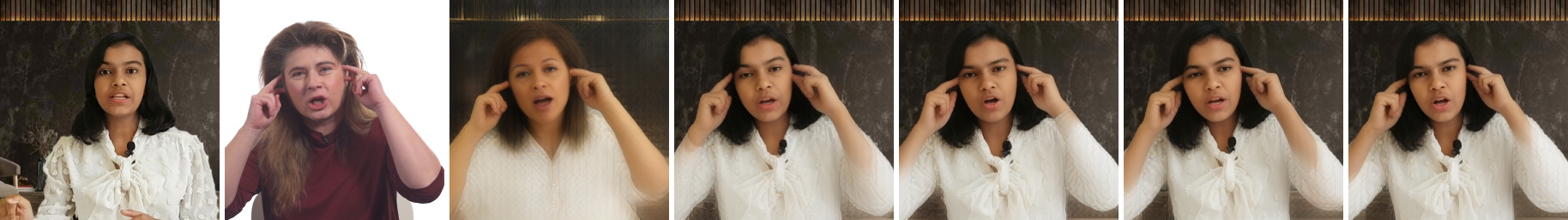}\\[0.5mm]
\includegraphics[width=\textwidth]{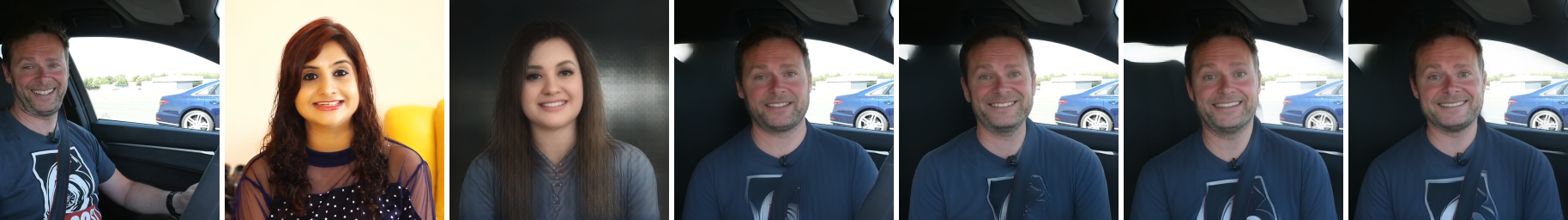}\\[0.5mm]
\includegraphics[width=\textwidth]{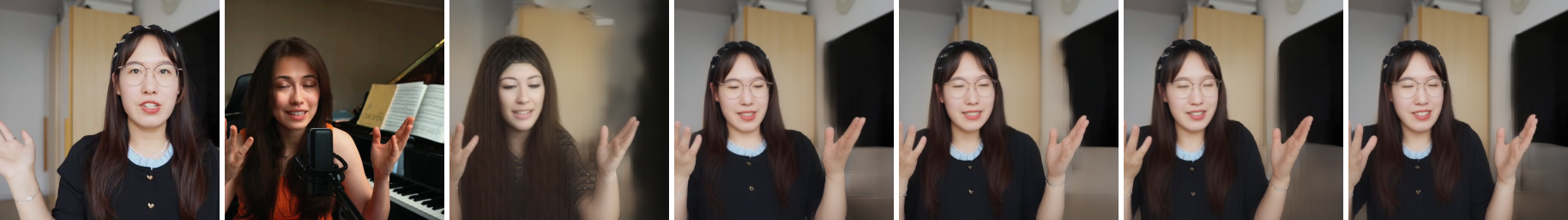}\\[0.5mm]
\includegraphics[width=\textwidth]{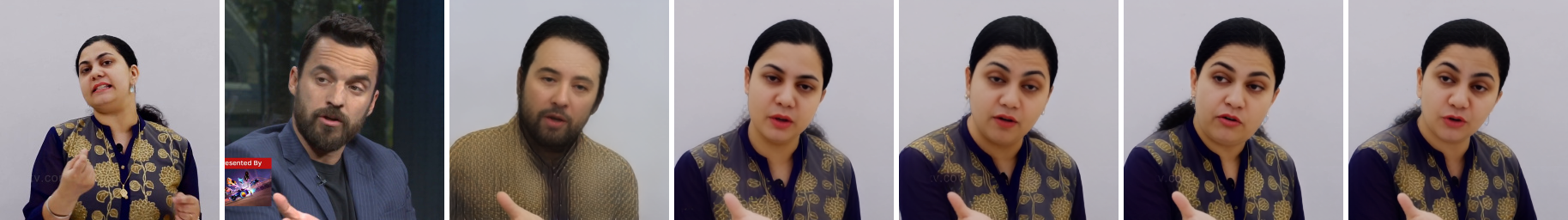}\\[0.5mm]
\includegraphics[width=\textwidth]{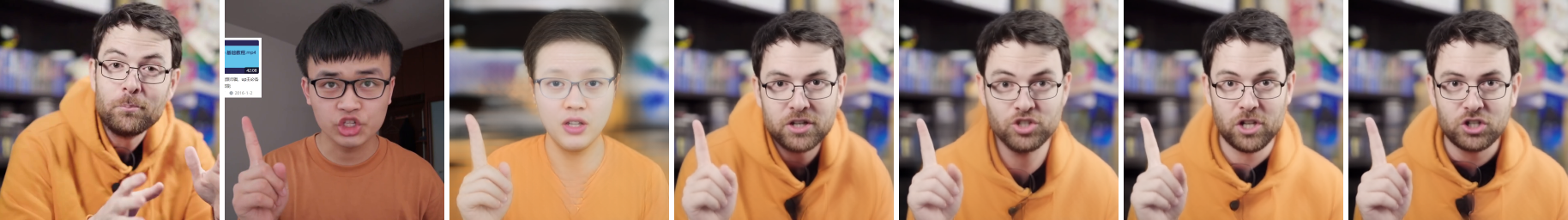}
\caption{\textbf{Qualitative ablation comparisons} on the upper-body cross-reenactment set.
Each row is a (reference, driver) pair; from left to right we show the reference image, the driver frame, four component ablations, and the full FUHA-C model.
Without the contrastive loss (third column), outputs take the driver's hair, gender, and face shape at low detail, but the reference's background and clothing; the other ablations stay close to FUHA-C.
See \autoref{tab:ablation_sep26} for matching numerical values.}
\label{fig:ablation_qual}
\end{figure*}

We provide the full FUHA-C ablation numbers in \autoref{tab:ablation_sep26} (self- and cross-reenactment) and qualitative samples in \autoref{fig:ablation_qual}.
All variants use the Cosmos tokenizer and are trained for 60 epochs.
The ReLU contrastive variant raises cross-ID to 0.730 but is worse than softplus on every self-reenactment metric and on every cross-reenactment motion error.

\input{table/ablation_sep26}

\section{Implementation Details}
\label{sec:suppl_impl}

\subsection{Architecture}
\label{sec:suppl_arch}

\ours\ uses a transformer-based encoder--decoder architecture with the components below.
All trainable transformers use Pre-LayerNorm, GELU activations, dropout 0.1, hidden dimension $d{=}512$, and feed-forward dimension $4d{=}2048$.

\parahead{Frozen feature extractors}
We use two frozen backbones:
(1)~a DINOv3 ViT-B/16~\citep{DINOv3} with 768-dimensional patch features at patch size 16, and
(2)~a Cosmos CV8$\times$8$\times$8 tokenizer~\citep{COSMOS} with a spatial compression factor of 8 and 16 latent channels.
FUHA-T replaces Cosmos with TAESD3. Both the visual encoder and the tokenizer are kept frozen throughout training.
Image-space losses backpropagate through the tokenizer decoder but do not update its weights.

\parahead{Feature fusion (appearance side only)}
The reference image's tokenizer features and DINOv3 features are independently projected to dimension $d$ via two-layer MLPs ($d_\text{in} \to d \to d$, GELU + LayerNorm) and summed at every spatial location, with a 2D sinusoidal positional encoding $\sigma_{2D}$ added before the appearance encoder.
Driver features use the same projection but are not fused with tokenizer features; the motion encoder reads DINOv3 features only.

\parahead{Appearance encoder}
A Q-Former~\citep{Blip2} with 12 transformer-decoder layers and 8 attention heads, attending with $N_a{=}1024$ learned query tokens.

\parahead{Motion encoder}
A Q-Former with 8 layers and 8 heads, attending with $N_m{=}2$ learned query tokens to driver DINOv3 features only.

\parahead{Motion decoder}
A Q-Former with 8 layers and 8 heads.
Its $L^2$ learned spatial queries (with $\sigma_{2D}$) attend to the motion tokens, producing a motion map matched to the tokenizer latent grid: $L{=}64$ at $512{\times}512$ input ($L{=}32$ at $256{\times}256$).

\parahead{Appearance decoder}
A transformer decoder with 16 layers and 8 heads.
Its queries are the motion map (with PE re-applied); its memory is the appearance embeddings $\mathbf{a}$.
The output is projected to the tokenizer latent dimension via a two-layer MLP and decoded by the frozen tokenizer decoder.

The transformer modules contain 189M trainable parameters: 51M in the appearance encoder, evaluated once per reference, and 138M in the modules evaluated for each driving frame. Including the frozen DINOv3 and Cosmos backbones, FUHA-C contains 380M parameters.

\subsection{Training}
\label{sec:suppl_training}

\parahead{Optimizer}
We use AdamW~\citep{loshchilov2019decoupled} with peak learning rate $5 \times 10^{-4}$ and no weight decay.
The learning rate follows a OneCycleLR schedule with linear warmup over 1 epoch and cosine annealing, starting and ending at $\text{lr}_\text{peak} / 10^4$.
Gradients are clipped to a maximum norm of 1.0.
We train for 60 epochs (approx.\ 3 days) on 64 H100 GPUs (8 nodes) with BF16 mixed precision and DDP.
Each GPU processes batches of 8 with 2 gradient accumulation steps, for an effective batch size of 1{,}024.

\parahead{Loss weights}
The objective combines the three losses defined in \cref{sec:objectives}:
\begin{equation}
    \mathcal{L} = \mathcal{L}_\text{rec}
    + \lambda_\text{tri}\mathcal{L}_\text{tri}
    + \lambda_\text{id}\mathcal{L}_\text{id},
    \label{eq:total_loss}
\end{equation}
which expands as
\begin{equation}
  \mathcal{L} = \lambda_\text{L1} \mathcal{L}_\text{L1}
  + \lambda_\text{LPIPS} \mathcal{L}_\text{LPIPS}
  + \lambda_\text{tri} \mathcal{L}_\text{tri}
  + \lambda_\text{id} \mathcal{L}_\text{id},
\end{equation}
with $\lambda_\text{L1}{=}1.0$, $\lambda_\text{LPIPS}{=}1.0$, $\lambda_\text{tri}{=}0.2$, and $\lambda_\text{id}{=}0.01$.

\parahead{Spatial weighting}
The image-space losses (L1 and LPIPS) are spatially weighted using per-pixel region masks.
The per-region weights are: background 0.1, foreground 0.2, face 1.0, eyes 10.0, mouth 5.0, hands 10.0.
Foreground masks are produced by BiRefNet~\citep{BiRefNet}; face/eyes/mouth/hands masks are computed on the fly from precomputed keypoints (\autoref{sec:suppl_data}).

\parahead{Contrastive loss}
We apply a triplet loss (\autoref{eq:triplet}) on the L2-normalized, flattened motion embedding with cosine distance $\delta$ and softplus activation (margin 0).
The normalization is applied only when computing the loss; the decoder receives the unnormalized embeddings.
The anchor is the motion embedding of the driver frame.
The positive is the motion embedding of a second view of the driver: a rendered keypoint image when face keypoints are available (the majority of samples), or a foreground-segmented crop otherwise (BiRefNet matting~\citep{BiRefNet} applied to a grayscale or RGB foreground).
The negative is the motion embedding of the reference frame, which has the same identity as the target output but a different pose.

\parahead{Identity loss}
We use the AdaFace~\citep{AdaFace} cosine distance on aligned 5-point landmark face crops, computed against the driver/target frame (pose-matched, same identity).
The loss is gated to samples for which face keypoints are available.

\parahead{Input dropout}
During training, we apply token-level dropout (rate 0.1) to both the driver and reference feature sequences before the motion and appearance encoders, respectively, to improve robustness.

\parahead{Data augmentation}
\label{sec:suppl_aug}
We apply photometric augmentations to the driver input only, each with probability 0.5: brightness ($\pm 0.3$), contrast ($\pm 0.3$), hue shift ($\pm 40^\circ$), saturation shift ($\pm 50$), value shift ($\pm 40$), CLAHE (clip limit 4.0, tile grid $8 \times 8$), Gaussian blur, and Gaussian noise.
Gamma correction (range $[0.8, 1.2]$) is applied with probability 0.3.
Driver, target, and reference share one random square crop (scale 0.9--1.0).

\section{Evaluation Details}
\label{sec:suppl_eval}

We evaluate \ours in three settings: face, upper body, and full body.
Each has its own test set and metric set.

\subsection{Metric Definitions}
\label{sec:suppl_metrics}

\parahead{Spatial (per-frame)}
\begin{itemize}[leftmargin=*,nosep]
  \item \textbf{L1} $\downarrow$: pixel mean absolute error.
  \item \textbf{SSIM} $\uparrow$: standard SSIM, averaged across frames.
  \item \textbf{PSNR} $\uparrow$: standard PSNR (full-body only).
  \item \textbf{LPIPS} $\downarrow$: LPIPS with AlexNet backbone~\citep{LPIPS}.
\end{itemize}

\parahead{Temporal}
\begin{itemize}[leftmargin=*,nosep]
  \item \textbf{FVD} $\downarrow$: Fr\'echet distance computed at the dataset level over 400-dim I3D features extracted in 16-frame chunks at $224\times224$ resolution (StyleGAN-V protocol~\citep{skorokhodov2022stylegan}). The real videos are the ground-truth targets for self-reenactment and the driving videos for cross-reenactment.
\end{itemize}

\parahead{Distribution}
\begin{itemize}[leftmargin=*,nosep]
  \item \textbf{FID} $\downarrow$: standard Inception-v3 FID, computed dataset-wide over all predicted and real frames (ground-truth targets for self-reenactment, driving frames for cross-reenactment).
\end{itemize}
In cross-reenactment, the real set shows the driver's identity, clothing, and background, so we read FID and FVD together with ID.
The 95\% intervals are paired bootstrap intervals of per-clip differences to FUHA-C (10{,}000 resamples of clips or pairs); they reflect clip sampling, not seed variance, and do not apply to the dataset-level FID and FVD.

\parahead{Identity and Face-Driven}
\begin{itemize}[leftmargin=*,nosep]
  \item \textbf{ID} $\uparrow$: ArcFace~\citep{ArcFace} identity similarity (\texttt{InsightFace} \texttt{buffalo\_l} embeddings), averaged across frames. For self-reenactment, the target for each predicted frame $\text{pred}_t$ is the driving frame at the same time step $\text{drv}_t$, following~\citet{LivePortrait}. For cross-reenactment, where no per-frame ground truth in the source identity exists, we use similarity against the static reference image, following prior reenactment evaluation~\citep{HunyuanPortrait,XNeMo,PersonaLive,XPortrait}.
  \item \textbf{AED} $\downarrow$: L1 distance between SMIRK~\citep{SMIRK} expression parameters (50-dim) of prediction and target (GT for self-reenactment; driving frame for cross-reenactment).
  \item \textbf{APD} $\downarrow$: L1 distance between SMIRK head-rotation parameters (3-dim axis-angle) of prediction and target.
\end{itemize}

\parahead{Body- and Hand-Pose (body settings only)}
We use the RTMW-x 133-keypoint wholebody estimator~\citep{RTMW} (\emph{performance} mode, $384\times288$ input).
A keypoint is included only if both GT and prediction detection confidences exceed 0.3.
Distance is raw pixel L2 after resizing predictions to GT resolution.
\begin{itemize}[leftmargin=*,nosep]
  \item \textbf{AKD-B} $\downarrow$: average pixel L2 over the 17 COCO body keypoints.
  \item \textbf{AKD-H} $\downarrow$: average pixel L2 over 42 hand keypoints (21 per hand).
\end{itemize}

\parahead{Face crop for body settings}
For AED in body settings (face is small in frame), we detect the GT face with InsightFace (largest bounding box, padded by $0.3$, cached so prediction and GT share the same crop), resize to $256\times256$, then run SMIRK. Frames with no GT face detection are excluded.
For ID, images are not pre-cropped: the standard InsightFace pipeline (detection $\to$ 5-keypoint similarity warp $\to$ $112\times112$ aligned face $\to$ ArcFace) is applied identically to prediction and target, with InsightFace handling face localization.

\parahead{Face protocol}
We use the TalkingHead-1KH~\citep{TalkingHead1KH} test set at $512\times512$, 128 frames per clip.
One of its test videos also appears in our training data; excluding its clips leaves the ranking on every per-clip face metric unchanged.
\emph{Self-reenactment (37 clips):} frame 0 is the reference; frames $1{:}128$ are the driver and ground truth. Metrics: L1, SSIM, LPIPS, ID, AED, APD, FID, FVD.
\emph{Cross-reenactment (100 pairs):} the first 100 FFHQ~\citep{StyleGAN} images serve as reference identities, each paired with a clip drawn at random, with replacement, from the 37 TalkingHead-1KH clips (33 distinct). Metrics: ID, AED, APD, FID, FVD (no pixel-paired metrics, since no ground truth exists).

\parahead{Upper-body protocol}
We use our upper-body benchmark with 96 self-reenactment clips and 96 cross-reenactment pairs, curated from public video sources following \autoref{sec:suppl_data}.
Test clips come from videos excluded from training.
As is common in web-scale video data, the same people, and occasionally the same footage, can recur in training videos.
\emph{Self-reenactment (96 clips):} L1, SSIM, LPIPS, ID, AED, AKD-B, AKD-H, FID, FVD.
\emph{Cross-reenactment (96 pairs):} each clip is the reference once and the driver once, never paired with itself. Metrics: ID, AED, AKD-B, AKD-H, FID, FVD.

\parahead{Full-body protocol}
We use the TikTok dance test set~\citep{jafarian2021learning} (10 clips) for full-body self-reenactment only; no cross-reenactment protocol is standardized for this benchmark.
Metrics: L1, SSIM, PSNR, LPIPS, ID, AED, AKD-B, AKD-H, FID, FVD.

\parahead{Resolution}
We evaluate face and upper-body at $512\times 512$, our training resolution, and full-body at the source resolution of the TikTok videos.

\parahead{Test-set availability}
The test sets of CyberHost~\citep{CyberHost}, OmniHuman-1~\citep{OmniHuman}, X-UniMotion~\citep{XUniMotion}, and Animate-X~\citep{AnimateX} are not public.
We therefore curate our upper-body benchmark (96 self-reenactment clips, 96 cross-reenactment pairs) and use the public TikTok dance test split~\citep{jafarian2021learning} (10 clips) for full-body, matching the protocol of DreamActor-M1~\citep{DreamActorM1}.

\section{Dataset Details}
\label{sec:suppl_data}

We describe the preprocessing details for our curated training dataset. The same pipeline produced the upper-body test set of \autoref{tab:upperbody}.

\parahead{Composition}
We aggregate raw video files from four public sources: HD-VILA-100M~\citep{HDVILA}, CelebV-Text~\citep{CelebVText}, TalkVid~\citep{TalkVid}, and SpeakerVid-5M~\citep{SpeakerVid}.
Of 430{,}989 source videos, 169{,}443 pass all filtering stages, yielding 1{,}056{,}941 clips of 10--20 seconds each and totaling 5{,}172.1~hours of footage.
Each clip carries per-frame body, face, and hand keypoint annotations.

\parahead{Preprocessing pipeline}
\label{sec:suppl_preproc}
\autoref{tab:preprocessing} summarizes the per-stage filtering applied to source videos. We additionally require $\geq$720p source resolution, apply $2\times$ bounding-box expansion around the detected person, and resample to $25$\,FPS.

\begin{table}[!ht]
\centering
\setlength{\tabcolsep}{6pt}
\renewcommand{\arraystretch}{1.05}
\small
\caption{\textbf{Training data processing pipeline.}}
\label{tab:preprocessing}
\begin{tabular}{@{}l l p{0.33\linewidth}@{}}
\toprule
Stage & Model used & Criterion / output \\
\midrule
Shot boundaries & TransNetV2~\citep{TransNetV2} & score threshold $0.2$; $10$--$20$\,s shots \\
Single-human detection & RTMW-x~\citep{RTMW} & conf $\geq 0.5$ on $5$ sampled frames; visible motion \\
Video-quality filter & DOVER~\citep{DOVER} & score $\geq 0.50$ \\
Static-background filter & InSPyReNet~\citep{InSPyReNet} & median background pixel std $\leq 1.0$ \\
Text-overlay filter & FAST-Tiny~\citep{FAST} & detected text area $\leq 2\%$ of frame \\
\midrule
Body keypoints & RTMPose-Body~\citep{RTMPose} & $17$ COCO-format keypoints \\
Face keypoints & SPIGA~\citep{SPIGA} & $98$ landmarks $+$ head pose \\
Hand keypoints & WiLoR-mini~\citep{WiLoR} & $21$ 3D keypoints per hand \\
\bottomrule
\end{tabular}
\end{table}

\section{Efficiency}
\label{sec:suppl_efficiency}

\autoref{fig:efficiency} reports single-GPU inference in eager mode, with most methods measured on an RTX Pro 6000; FollowYE uses a Titan V and Champ an RTX 4090.
FUHA-C/T use BF16 mixed precision.
We report the median of three passes over 128 driving frames (125 for UniAnimate-DiT, which requires $4k{+}1$ frames) at $512^2$, including reference and motion encoding, image generation, and the baselines' pose, landmark, or guidance extraction; model loading, video decoding, and file I/O are excluded.
The RTX 4090 speeds in \cref{sec:efficiency} use the four TorchScript-exported modules in FP16, frame by frame (batch size 1) over 128 synthetic $512^2$ driving frames, with the appearance encoded once inside the timed region and host-to-device copies excluded; we report the median of three passes after two warm-up passes.
Peak VRAM on the RTX 4090 is 3.6\,GB (FUHA-C) and 2.6\,GB (FUHA-T).
Peak VRAM is the PyTorch allocator peak after one warm-up pass (FollowYE and Champ: without warm-up, on their GPUs).
The diffusion baselines use 4 (PersonaLive), 20 (Wan-Animate, Champ), 25 (HunyuanPortrait, MimicMotion), 30 (Moore-AA, X-Portrait, FollowYE), 35 (X-NeMo), and 50 (UniAnimate-DiT, with TeaCache) denoising steps.

\section{Limitations}
\label{sec:suppl_limitations}
\ours renders each frame independently, without temporal modeling or a generative prior; this favors speed and statelessness, and on full-body the best diffusion baselines are temporally smoother (\autoref{sec:suppl_tlp}).
Fine detail such as hair and fabric, and regions hidden in the reference, are softer than with diffusion-based methods, the best of which obtain lower FID and FVD in cross-reenactment and on full-body.
Training relies on keypoints from offline estimators and inherits their errors.
Our data contains no non-human characters, which \ours animates less reliably.
A video diffusion decoder driven by the same frozen motion embedding (FUHA-W) closes the upper-body cross-FVD gap but runs far from real time (\autoref{sec:suppl_diffusion}).
Future work includes a faster, temporally aware decoder for the motion embedding, and driving the embedding from audio.

%% file: section/retrieval_sep26.tex
\section{Motion Representation}
\label{sec:suppl_motion}
We probe the motion embedding in three ways: nearest-neighbor retrieval, interpolation, and conditioning a different decoder.

\subsection{Motion Retrieval}
\label{sec:suppl_retrieval}

We probe the motion representation independently of the animation decoder by retrieving a nearest neighbor for each of 782 test frames from 10{,}442 validation frames. The query and gallery sets are disjoint by person: overlapping identities were removed from the gallery by ArcFace matching, followed by a manual check of the closest remaining matches. FUHA-C represents each frame with its motion embedding (1{,}024 dimensions). The comparison encoders use their learned global embedding per frame (CLS token or pooled output), and retrieval uses cosine similarity.

In \autoref{tab:retrieval_sep26}, body and hand keypoint distances are measured at $512^2$, and head-pose error is measured in degrees. Hand retrieval uses 475 queries with visible hands and restricts the gallery to frames showing the query's visible hands. Same-person retrieval uses the gallery before identity removal (15{,}122 frames, 796 videos); a top-1 frame counts as the same person if its ArcFace similarity to the query reaches 0.25, the 99.9th percentile over random cross-video frame pairs. Rates are over all 782 queries; 296 have such a frame in this gallery.

FUHA-C has lower body-pose error than all ten learned global embeddings. Its hand-pose error is comparable to the strongest of these embeddings, while head-pose error is similar to Sapiens2 and DINOv3. On same-person retrieval, FUHA-C retrieves the query's person in 0.1\% of cases, compared with 3.2--11.5\% for the learned global embeddings. The motion embedding thus retains body pose but little identity.

\input{table/retrieval_sep26}

\begin{figure*}[!t]
\centering
\includegraphics[width=\textwidth]{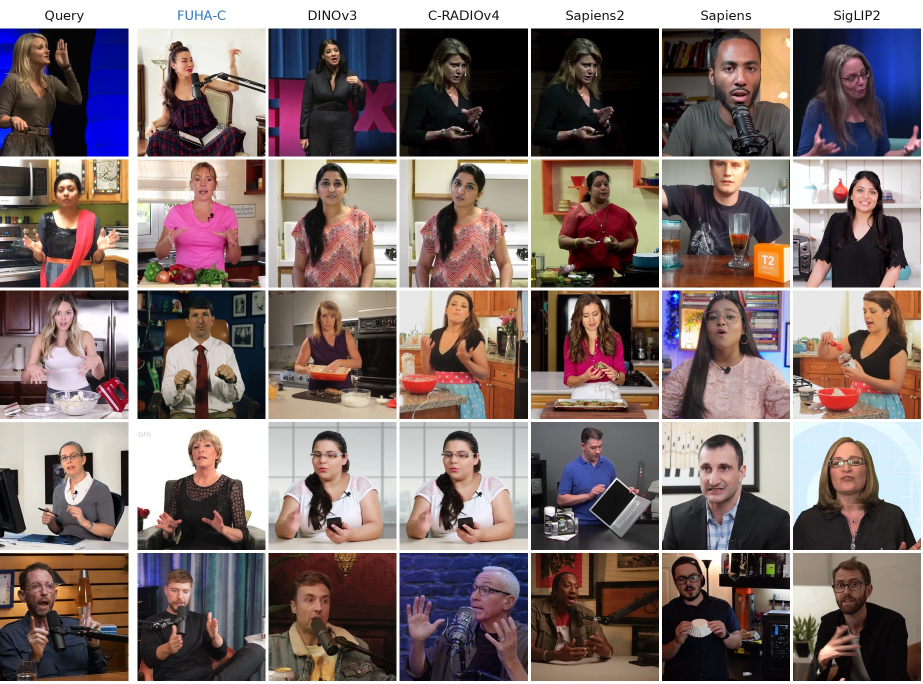}
\caption{\textbf{Motion retrieval across identities.} Selected queries (left) and their nearest neighbors from FUHA-C and five vision-encoder baselines. The gallery excludes the query's identity. Baseline variants are DINOv3 ViT-B/16, C-RADIOv4-SO400M, Sapiens2-0.1B, Sapiens-0.3B, and SigLIP2-base-p16-224, from left to right.}
\label{fig:retrieval_examples}
\end{figure*}

\autoref{fig:retrieval_examples} shows selected retrievals from the person-disjoint gallery.

\subsection{Motion Latent-Space Walks}
\label{sec:suppl_walks}
\begin{figure}[!t]
\centering
\includegraphics[width=0.6\linewidth]{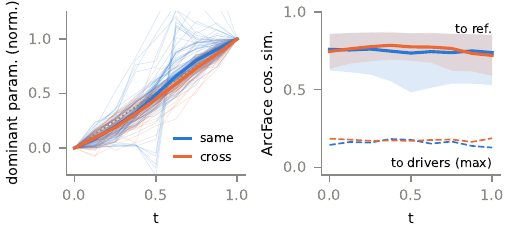}
\caption{\textbf{Walk measurements} over 25 walks and six references. Left: dominant SMIRK parameter, normalized by the decoded endpoints (thin: walk and reference; thick: median; dotted: linear). Right: ArcFace similarity to the reference (solid: median; band: 10th--90th percentile) and to either driver (dashed: maximum). Same: both drivers from one clip; cross: from two people.}
\label{fig:walk_curves}
\end{figure}

\begin{figure*}[p]
\centering
\includegraphics[width=\textwidth]{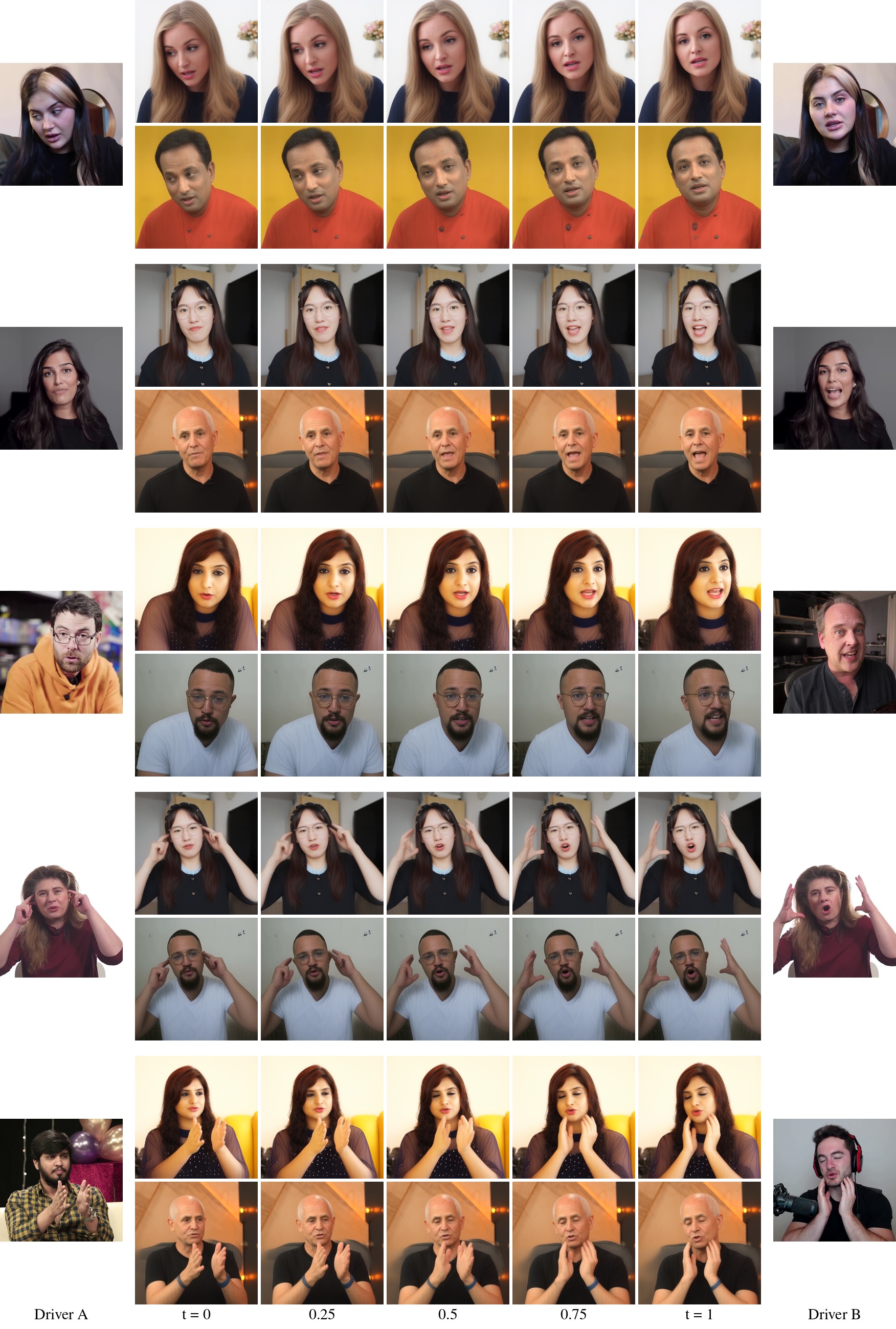}
\caption{\textbf{Motion latent-space walks.} We linearly interpolate between the FUHA-C motion embeddings of two driver frames (left and right) and decode each step on two fixed references. The top three walks change head pose and expression; the bottom two move the hands. Walks 3 and 5 interpolate between two different people. Motion changes smoothly while the reference identity is preserved.}
\label{fig:latent_walks}
\end{figure*}

Using FUHA-C, we linearly interpolate between the motion embeddings of two driver frames, $\mathbf{m}(t) = (1-t)\,\mathbf{m}_A + t\,\mathbf{m}_B$, and decode each step on a fixed reference (\autoref{fig:latent_walks}).
Over 25 head-pose and expression walks (16 within a clip, 9 across people) and six references, the dominant SMIRK~\citep{SMIRK} parameter typically changes monotonically and almost linearly in $t$ (median Spearman $\rho = 1.00$, $R^2 = 0.99$; \autoref{fig:walk_curves}).
Each walk pairs two test frames that differ mainly in one factor (head yaw, pitch, roll, torso lean, mouth, eyes, or smile; two cross-person walks mix pose and expression); $\rho$ and $R^2$ are medians over walks of per-walk medians over the six references.
ArcFace~\citep{ArcFace} similarity to the reference stays flat (median 0.75), and similarity to either driver stays at or below 0.19.
The motion embedding space is thus smooth and shared across people: straight lines between two embeddings decode to gradual, near-linear motion, cross-person walks behave like same-clip ones, and the rendered identity stays with the reference throughout.
The bottom two walks in \autoref{fig:latent_walks} also articulate the hands.

%% file: table/retrieval_sep26.tex
\begin{table*}[!ht]
\centering
\caption{\textbf{Motion retrieval with learned global embeddings.} Body/hand keypoint errors (AKD-B/H) are in pixels at $512^2$; head-pose error is in degrees. Pose scores use a person-disjoint gallery; same-person retrieval is measured separately with shared identities retained. Lower is better. AKD-H uses 475 queries with visible hands.}
\label{tab:retrieval_sep26}
\setlength{\tabcolsep}{4pt}
\small
\begin{tabular}{lcccc}
\toprule
Encoder & AKD-B $\downarrow$ & AKD-H $\downarrow$ & Head pose $\downarrow$ & Same person $\downarrow$ \\
 & (px) & (px) & (degrees) & (\%) \\
\midrule
Random & 110.8 & 166.2 & 10.61 & 0.0 \\
SigLIP2 base-p16-224 & 91.0 & 157.1 & 10.23 & 10.4 \\
SigLIP2 so400m-p16-512 & 91.3 & 152.4 & 10.09 & 11.3 \\
Sapiens 0.3B & 98.9 & 167.9 & 10.22 & 3.2 \\
Sapiens 1B & 102.3 & 162.1 & 10.35 & 3.8 \\
Sapiens2 0.1B & 85.5 & 143.8 & 9.09 & 9.5 \\
Sapiens2 1B & 91.4 & 153.2 & 9.24 & 8.6 \\
C-RADIOv4 SO400M & 88.7 & 148.1 & 9.90 & 11.5 \\
C-RADIOv4 H & 88.2 & 149.2 & 9.76 & 11.5 \\
DINOv3 ViT-B/16 & 84.9 & 142.6 & 9.19 & 9.5 \\
DINOv3 ViT-L/16 & 89.1 & 146.1 & 9.18 & 9.6 \\
\midrule
FUHA-C & 78.1 & 140.4 & 9.18 & 0.1 \\
\bottomrule
\end{tabular}
\end{table*}

%% file: table/wan_dit_suppl.tex
\begin{table*}[t]
\centering
\caption{\textbf{Animation with a diffusion decoder.} FUHA-W uses the FUHA-C motion embedding to condition Wan. Baseline results are reproduced from Tables~\ref{tab:upperbody}--\ref{tab:fullbody}. Cells are ranked within each setting and region (\colorbox{rankbest}{1st}, \colorbox{ranksecond}{2nd}, \colorbox{rankthird}{3rd}).}
\label{tab:wan}
\setlength{\tabcolsep}{1.5pt}
\renewcommand{\arraystretch}{1.15}
\footnotesize
\begin{tabular*}{\textwidth}{@{\extracolsep{\fill}}llccccccccc@{}}
\toprule
\multicolumn{11}{l}{\textbf{(a) Upper-body}} \\
\midrule
Setting & Model & L1 $\downarrow$ & SSIM $\uparrow$ & LPIPS $\downarrow$ & ID $\uparrow$ & AED $\downarrow$ & AKD-B $\downarrow$ & AKD-H $\downarrow$ & FID $\downarrow$ & FVD $\downarrow$ \\
\midrule
Self & Wan-Animate & \cellcolor{rankthird}0.0643 & 0.712 & 0.219 & 0.668 & \cellcolor{ranksecond}0.467 & \cellcolor{rankthird}37.6 & \cellcolor{rankthird}70.5 & 24.4 & 133.4 \\
 & UniAnimate-DiT & 0.0717 & \cellcolor{rankthird}0.724 & \cellcolor{rankthird}0.201 & \cellcolor{rankthird}0.694 & 0.618 & 46.6 & 86.6 & \cellcolor{rankthird}20.3 & \cellcolor{rankthird}66.0 \\
 & FUHA-C & \cellcolor{rankbest}0.0287 & \cellcolor{rankbest}0.838 & \cellcolor{rankbest}0.106 & \cellcolor{rankbest}0.820 & \cellcolor{rankbest}0.315 & \cellcolor{rankbest}27.1 & \cellcolor{rankbest}48.5 & \cellcolor{rankbest}15.2 & \cellcolor{ranksecond}38.9 \\
 & FUHA-W & \cellcolor{ranksecond}0.0413 & \cellcolor{ranksecond}0.781 & \cellcolor{ranksecond}0.140 & \cellcolor{ranksecond}0.781 & \cellcolor{rankthird}0.472 & \cellcolor{ranksecond}32.8 & \cellcolor{ranksecond}56.6 & \cellcolor{ranksecond}18.4 & \cellcolor{rankbest}27.9 \\
\midrule
Cross & Wan-Animate & -- & -- & -- & 0.489 & \cellcolor{ranksecond}0.765 & \cellcolor{rankbest}50.8 & \cellcolor{ranksecond}93.4 & 57.5 & 207.4 \\
 & UniAnimate-DiT & -- & -- & -- & \cellcolor{rankthird}0.630 & 0.850 & \cellcolor{rankthird}55.7 & 97.2 & \cellcolor{rankbest}50.2 & \cellcolor{ranksecond}145.8 \\
 & FUHA-C & -- & -- & -- & \cellcolor{rankbest}0.701 & \cellcolor{rankbest}0.686 & \cellcolor{ranksecond}51.5 & \cellcolor{rankbest}88.8 & \cellcolor{ranksecond}55.9 & \cellcolor{rankthird}171.7 \\
 & FUHA-W & -- & -- & -- & \cellcolor{ranksecond}0.676 & \cellcolor{rankthird}0.827 & 61.4 & \cellcolor{rankthird}94.8 & \cellcolor{rankthird}56.4 & \cellcolor{rankbest}125.0 \\
\bottomrule
\end{tabular*}
\par\medskip
\begin{tabular*}{\textwidth}{@{\extracolsep{\fill}}llcccccccc@{}}
\toprule
\multicolumn{10}{l}{\textbf{(b) Face}} \\
\midrule
Setting & Model & L1 $\downarrow$ & SSIM $\uparrow$ & LPIPS $\downarrow$ & ID $\uparrow$ & AED $\downarrow$ & APD $\downarrow$ & FID $\downarrow$ & FVD $\downarrow$ \\
\midrule
Self & PersonaLive & \cellcolor{rankthird}0.0721 & \cellcolor{rankthird}0.698 & \cellcolor{rankthird}0.239 & \cellcolor{ranksecond}0.870 & \cellcolor{ranksecond}0.323 & \cellcolor{ranksecond}0.0171 & \cellcolor{rankbest}32.6 & \cellcolor{rankthird}210.1 \\
 & FUHA-C & \cellcolor{rankbest}0.0526 & \cellcolor{rankbest}0.782 & \cellcolor{rankbest}0.185 & \cellcolor{rankbest}0.876 & \cellcolor{rankbest}0.293 & \cellcolor{rankbest}0.0131 & \cellcolor{ranksecond}35.0 & \cellcolor{rankbest}153.8 \\
 & FUHA-W & \cellcolor{ranksecond}0.0685 & \cellcolor{ranksecond}0.724 & \cellcolor{ranksecond}0.234 & \cellcolor{rankthird}0.855 & \cellcolor{rankthird}0.429 & \cellcolor{rankthird}0.0297 & \cellcolor{rankthird}42.0 & \cellcolor{ranksecond}189.2 \\
\midrule
Cross & PersonaLive & -- & -- & -- & \cellcolor{ranksecond}0.698 & \cellcolor{ranksecond}0.811 & \cellcolor{ranksecond}0.0365 & \cellcolor{rankbest}154.9 & \cellcolor{rankbest}561.3 \\
 & FUHA-C & -- & -- & -- & \cellcolor{rankthird}0.654 & \cellcolor{rankbest}0.750 & \cellcolor{rankbest}0.0321 & \cellcolor{rankthird}158.2 & \cellcolor{rankthird}645.3 \\
 & FUHA-W & -- & -- & -- & \cellcolor{rankbest}0.763 & \cellcolor{rankthird}0.929 & \cellcolor{rankthird}0.0651 & \cellcolor{ranksecond}157.1 & \cellcolor{ranksecond}581.4 \\
\bottomrule
\end{tabular*}
\par\medskip
\begin{tabular*}{\textwidth}{@{\extracolsep{\fill}}llcccccccccc@{}}
\toprule
\multicolumn{12}{l}{\textbf{(c) Full-body}} \\
\midrule
Setting & Model & L1 $\downarrow$ & SSIM $\uparrow$ & PSNR $\uparrow$ & LPIPS $\downarrow$ & ID $\uparrow$ & AED $\downarrow$ & AKD-B $\downarrow$ & AKD-H $\downarrow$ & FID $\downarrow$ & FVD $\downarrow$ \\
\midrule
Self & Wan-Animate & \cellcolor{rankthird}0.0815 & 0.768 & \cellcolor{rankthird}17.76 & 0.310 & 0.427 & \cellcolor{ranksecond}0.705 & \cellcolor{rankthird}46.8 & \cellcolor{rankthird}84.5 & \cellcolor{ranksecond}36.1 & \cellcolor{rankbest}168.2 \\
 & UniAnimate-DiT & \cellcolor{ranksecond}0.0672 & \cellcolor{ranksecond}0.818 & \cellcolor{ranksecond}19.11 & \cellcolor{ranksecond}0.247 & \cellcolor{ranksecond}0.458 & \cellcolor{ranksecond}0.705 & \cellcolor{ranksecond}44.9 & \cellcolor{ranksecond}76.4 & \cellcolor{rankbest}25.9 & \cellcolor{ranksecond}170.8 \\
 & FUHA-C & \cellcolor{rankbest}0.0559 & \cellcolor{rankbest}0.852 & \cellcolor{rankbest}20.50 & \cellcolor{rankbest}0.236 & \cellcolor{rankbest}0.519 & \cellcolor{rankbest}0.605 & \cellcolor{rankbest}44.2 & \cellcolor{rankbest}69.5 & \cellcolor{rankthird}37.9 & \cellcolor{rankthird}230.5 \\
 & FUHA-W & 0.0842 & \cellcolor{rankthird}0.795 & 16.98 & \cellcolor{rankthird}0.309 & \cellcolor{rankthird}0.438 & \cellcolor{rankthird}0.751 & 58.7 & 92.2 & 54.3 & 260.8 \\
\bottomrule
\end{tabular*}
\par\medskip
\end{table*}

%% file: table/tlp_self.tex
\begin{table}[!ht]
\centering
\caption{\textbf{Temporal consistency}: self-reenactment tLP~$\downarrow$ ($\times 100$). -- marks methods not evaluated on a benchmark. Cells are shaded by per-column rank: \colorbox{rankbest}{1st}, \colorbox{ranksecond}{2nd}, and \colorbox{rankthird}{3rd}.}
\label{tab:tlp}
\small
\begin{tabular}{lccc}
\toprule
Method & Face & Upper-body & Full-body \\
\midrule
LivePortrait & 1.96 & -- & -- \\
X-Portrait & 2.60 & -- & -- \\
FollowYE & \cellcolor{ranksecond}1.90 & -- & -- \\
X-NeMo & 1.97 & -- & -- \\
HunyuanPortrait & 2.01 & -- & -- \\
PersonaLive & 2.47 & -- & -- \\
\midrule
Moore-AA & -- & 2.61 & \cellcolor{rankbest}2.33 \\
Champ & -- & 3.21 & 3.77 \\
MimicMotion & -- & \cellcolor{rankthird}0.77 & \cellcolor{rankthird}2.62 \\
Wan-Animate & -- & 2.59 & 4.41 \\
UniAnimate-DiT & -- & 1.01 & \cellcolor{ranksecond}2.50 \\
Animate-X & -- & -- & 3.47 \\
\midrule
FUHA-C & \cellcolor{rankbest}1.87 & \cellcolor{rankbest}0.47 & 3.22 \\
FUHA-T & \cellcolor{rankthird}1.93 & \cellcolor{ranksecond}0.56 & 3.45 \\
\bottomrule
\end{tabular}
\end{table}

%% file: table/ablation_sep26.tex
\begin{table*}[!ht]
\centering
\caption{\textbf{FUHA-C ablations} on the upper-body set at $512^2$.
(\colorbox{rankbest}{1st}, \colorbox{ranksecond}{2nd}, \colorbox{rankthird}{3rd}).}
\label{tab:ablation_sep26}
\setlength{\tabcolsep}{4pt}
\scriptsize
\resizebox{\textwidth}{!}{%
\begin{tabular}{lcccccccccc}
\toprule
& \multicolumn{4}{c}{Self-Reenactment} & \multicolumn{6}{c}{Cross-Reenactment} \\
\cmidrule(lr){2-5} \cmidrule(lr){6-11}
Variant & LPIPS $\downarrow$ & ID $\uparrow$ & FID $\downarrow$ & FVD $\downarrow$
& ID $\uparrow$ & AED $\downarrow$ & AKD-B $\downarrow$ & AKD-H $\downarrow$
& FID $\downarrow$ & FVD $\downarrow$ \\
\midrule
1 motion token & 0.351 & 0.134 & 78.8 & 258.5 & 0.088 & 0.724 & 57.8 & 92.6 & 88.0 & 310.8 \\
2 motion tokens & \cellcolor{rankbest}0.106 & \cellcolor{rankbest}0.820 & \cellcolor{rankbest}15.2 & \cellcolor{rankbest}38.9 & 0.701 & 0.686 & \cellcolor{ranksecond}51.5 & 88.8 & \cellcolor{rankthird}55.9 & \cellcolor{ranksecond}171.7 \\
4 motion tokens & 0.127 & \cellcolor{rankthird}0.812 & 17.7 & 63.1 & \cellcolor{rankbest}0.733 & 0.761 & 58.9 & 91.6 & \cellcolor{ranksecond}55.7 & 201.9 \\
\midrule
ReLU contrastive & 0.125 & \cellcolor{ranksecond}0.813 & \cellcolor{rankthird}16.8 & 54.6 & \cellcolor{ranksecond}0.730 & 0.709 & 55.4 & 91.9 & \cellcolor{rankbest}55.6 & \cellcolor{rankthird}189.2 \\
w/o DINO in appearance & 0.125 & 0.811 & 18.3 & 59.4 & \cellcolor{rankthird}0.712 & 0.732 & 55.8 & 91.1 & 56.1 & 202.4 \\
w/o contrastive loss & 0.340 & 0.199 & 88.8 & 263.9 & 0.057 & \cellcolor{rankbest}0.606 & \cellcolor{rankbest}47.8 & \cellcolor{rankbest}80.3 & 101.1 & 325.9 \\
w/o 2D motion decoder & \cellcolor{rankthird}0.116 & 0.799 & 16.9 & \cellcolor{rankthird}49.2 & 0.689 & \cellcolor{rankthird}0.685 & \cellcolor{rankthird}51.9 & \cellcolor{rankthird}86.9 & 57.6 & 193.0 \\
w/o ID loss & \cellcolor{ranksecond}0.110 & 0.764 & \cellcolor{ranksecond}15.6 & \cellcolor{ranksecond}41.5 & 0.568 & \cellcolor{ranksecond}0.671 & 52.1 & \cellcolor{ranksecond}86.5 & 56.8 & \cellcolor{rankbest}170.6 \\
FUHA-C & \cellcolor{rankbest}0.106 & \cellcolor{rankbest}0.820 & \cellcolor{rankbest}15.2 & \cellcolor{rankbest}38.9 & 0.701 & 0.686 & \cellcolor{ranksecond}51.5 & 88.8 & \cellcolor{rankthird}55.9 & \cellcolor{ranksecond}171.7 \\
\bottomrule
\end{tabular}%
}
\end{table*}